\documentclass[a4paper,fleqn]{cas-dc}

\usepackage{amssymb}
\usepackage{amsmath}
\usepackage{array}
\usepackage{tabularx}
\usepackage[numbers,sort&compress]{natbib}  %
\usepackage{ragged2e}
\usepackage{float}
\usepackage{placeins}
\usepackage{multirow}
\usepackage{makecell}
\usepackage{booktabs}
\usepackage{longtable}
\usepackage{CJKutf8}
\usepackage{natbib}
\usepackage{footmisc}
\usepackage{geometry}
\usepackage{cleveref}

\def\tsc#1{\csdef{#1}{\textsc{\lowercase{#1}}\xspace}}
\tsc{WGM}
\tsc{QE}

\cortext[1]{Corresponding author}
\fntext[fn1]{\& These authors contributed equally to this work.}

\begin{document}

\let\WriteBookmarks\relax
\def\floatpagepagefraction{1}
\def\textpagefraction{.001}

\shorttitle{}    

\shortauthors{}  

\title [mode = title]{Embodied Intelligence: The Key to Unblocking Generalized Artificial Intelligence}  

\author[1]{Jinhao Jiang\textsuperscript{\&}}
\ead{2022215021@mail.hfut.edu.cn}

\author[2]{Changlin Chen\textsuperscript{\&}}
\ead{changlinchen@mail.ustc.edu.cn}

\author[1]{Shile Feng}
\ead{2024215125@mail.hfut.edu.cn}

\author[1]{Wanru Geng}
\ead{2022211618@mail.hfut.edu.cn}

\author[1]{Zesheng Zhou}
\ead{2023211500@mail.hfut.edu.cn}

\author[3]{Ni Wang}
\ead{niwang.fr@gmail.com}

\author[1]{Shuai Li}
\ead{2022216057@mail.hfut.edu.cn}

\author[4]{FengQi Cui}
\ead{fengqi_cui@mail.ustc.edu.cn}

\author[2]{Erbao Dong\textsuperscript{*}}
\ead[cor1]{ebdong@ustc.edu.cn}
\cortext[cor1]{Corresponding author}

\affiliation[1]{organization={Hefei University of Technology},
            addressline={}, 
            city={Hefei},
            postcode={230601}, 
            state={Anhui Province},
            country={China}}

\affiliation[2]{organization={University of Science and Technology of China},
            department={Institute of Humanoid Robots, Department of Precision Machinery and Precision Instrumentation}, 
            city={Hefei},
            postcode={230026}, 
            state={Anhui Province},
            country={China}} 

\affiliation[3]{organization={Amazon Development Center Germany GmbH}, 
            city={Berlin},
            postcode={10117}, 
            country={Germany}} 

\affiliation[4]{organization={University of Science and Technology of China},
            department={Institute of Advanced Technology}, 
            city={Hefei},
            postcode={230026}, 
            country={China}}
            
\begin{abstract}
The ultimate goal of artificial intelligence (AI) is to achieve Artificial General Intelligence (AGI). Embodied Artificial Intelligence (EAI), which involves intelligent systems with physical presence and real-time interaction with the environment, has emerged as a key research direction in pursuit of AGI. While advancements in deep learning, reinforcement learning, large-scale language models, and multimodal technologies have significantly contributed to the progress of EAI, most existing reviews focus on specific technologies or applications. A systematic overview, particularly one that explores the direct connection between EAI and AGI, remains scarce. This paper examines EAI as a foundational approach to AGI, systematically analyzing its four core modules: perception, intelligent decision-making, action, and feedback. We provide a detailed discussion of how each module contributes to the six core principles of AGI. Additionally, we discuss future trends, challenges, and research directions in EAI, emphasizing its potential as a cornerstone for AGI development. Our findings suggest that EAI’s integration of dynamic learning and real-world interaction is essential for bridging the gap between narrow AI and AGI.
\end{abstract}


\maketitle

\section{Introduction}

Artificial General Intelligence (AGI), regarded as the ultimate goal of AI research, refers to a system capable of achieving human-level intelligence. However, achieving AGI remains a significant challenge. Traditional symbolic AI models, which rely on logical reasoning and computational power, often overlook the importance of real-time interaction with the environment, limiting their effectiveness in complex real-world scenarios. Although deep learning and related technologies have improved the performance of intelligent systems, AGI development still faces challenges such as limited task generalization, high computational demands, and difficulties in integrating multimodal information\cite{dou2023towards}.
\par
Embodied intelligence (EAI) is defined as a system in which an agent perceives, learns and makes decisions through the interaction between its body and its environment\cite{2021Roy}. It offers a promising approach to addressing these limitations. Unlike purely computational AI models, EAI emphasizes intelligence that arises through real-time interaction with the environment. By integrating dynamic perception and adaptive behavior, EAI enables goal-oriented decision-making, making it one of the most viable pathways toward AGI\cite{grossberg2020path}. In recent years, EAI has demonstrated significant potential in soft robotics, smart homes, and other applications\cite{cianchetti2021embodied}. Advances in deep learning, reinforcement learning, and multimodal technologies have further accelerated its progress\cite{summaira2021recent}. Beyond practical applications, EAI is increasingly viewed as a key enabler of AGI, drawing widespread attention and significant industry investment\cite{2024Zhou}. Despite its potential, most existing studies focus on specific technologies or application areas rather than providing a comprehensive framework linking EAI to AGI. Notably, a joint review by Pengcheng Laboratory and HCP Laboratory of Sun Yat-sen University\cite{2024Liu} is among the most comprehensive analyses of EAI, yet it does not explore its direct relationship with AGI in depth.
\par
This paper addresses this gap by systematically examining EAI as a foundational approach to AGI. First of all, this paper innovatively divides the technical solutions of EI into sub-modular architecture and end-to-end architecture. This paper makes an in-depth technical deconstruction of the emerging end-to-end architecture. For modular architecture, we analyze its four core modules—perception, intelligent decision-making, action, and feedback—exploring how each contributes to AGI’s six fundamental principles. First, we review the development of EAI and its fundamental concepts. Next, we examine the functions and interactions of its core modules. We then discuss future research directions and challenges, providing a structured perspective on how EAI can accelerate AGI development.
\par
The remainder of this paper is organized as follows: Section 2 explores the evolution of EAI, Section 3 introduces the technical route of embodied intelligence and gives an overview of the end-to-end system framework, Section 4 discusses modularity analysis of sub-modular frameworks, section 4 discusses future trends and challenges, and Section 5 summarizes this paper.

\section{Conception of embodied intelligence}
The rise of embodied intelligence is closely linked to AGI, the ultimate goal of AI\cite{bariah2024ai}. AGI aims to achieve an intelligent system capable of learning autonomously, adapting to its environment, and performing multiple tasks\cite{1950Turing}. Since its introduction in the mid-20th century, AGI has remained a central topic in artificial intelligence research. However, brain-centric or computationally driven approaches often overlook how intelligence interacts with its physical embodiment and environment, limiting their applicability in complex real-world scenarios\cite{arslan2024artificial}. This limitation has led researchers to rethink the nature of intelligence, taking inspiration from biological intelligence and behavior\cite{floreano2008bio}.

\subsection{The Origins of AGI and Early Challenges}
The concept of AGI was first conceptualized in the 1950s. In 2002, Peter Voss et al. introduced the term "General Artificial Intelligence," defining it as an AI system capable of learning in real time, operating autonomously, and accomplishing any cognitive task with minimal resources\cite{2012Adams}. Some scholars speculate that if AGI systems are realized, their powerful optimization capabilities will enable them to improve themselves recursively and eventually surpass human intelligence by several orders of magnitude\cite{2021Silver}. Despite significant interest in AGI, multiple attempts to achieve it—such as the General Problem Solver\cite{2021Garvey}, the Fifth Generation Computer Systems Initiative \cite{1983Moto-Oka}, and DARPA's Strategic Computing Initiative\cite{2002Roland}—were unsuccessful. As technical challenges emerged, researchers shifted their focus toward developing domain-specific AI systems.

\subsection{The Shift from Narrow AI (ANI) to AGI}
In 2005, Ray Kurzweil introduced the term "narrow AI" (ANI) \cite{2014Goertzel} to differentiate it from AGI, emphasizing ANI's focus on specialized tasks while AGI aims for broad generalization capabilities (see Table~\ref{tab1}).
\par
Since the 21st century, breakthroughs in deep learning \cite{2020Sejnowski} and the powerful language comprehension capabilities of large language models \cite{2024Ge} have driven frequent mentions of generalized AGI and triggered a boom in AI development. Among the various paths to realize AGI, EA is considered the most promising one. EAI is able to interact with the environment in real time through sensing and interaction\cite{2004Pfeifer}. This concept traces back to foundational work by Alan Turing, who posited in his 1948 Intelligent Machinery report that true machine intelligence must develop through embodied physical engagement with the world. In contrast to ANI systems constrained by fixed programming architectures, Turing's framework established AGI as requiring adaptive environmental coupling — an epistemic shift later formalized by Brooks' subsumption architecture\cite{1991Brooks}. These embodied cognition principles continue to inform contemporary EAI research paradigms focused on sensorimotor contingency learning.

\begin{table}[H]
\fontfamily{ptm}\selectfont
\centering
\caption{Generalized AI vs. Narrow AI}
\begin{tabular}{ p{1.5cm} p{2.6cm} p{2.6cm}}\toprule
 \hline
  Feature & AGI  & ANI   \\
 \hline
  Target & To be as intelligent as a human being & Used only to address specific problems in specific areas \\
  \hline
 Ability to handle tasks & Multi-tasking,multi-domain & Focus on specific tasks\\ 
 \hline
 Operational features & Self-learning through interaction with the environ-ment & Runs through a fixed programming framework \\
 \hline
 Areas of app-lication & Medical, Robotics, etc & Voice assistants,image recognition,etc \\
 \hline
\end{tabular}
\label{tab1}
\end{table}

\subsection{The Emergence of embodied intelligence and its Core Theories}
The work of Alan Turing advanced research based on “underlying intelligence”, i.e., the gradual accumulation of information through simple perceptual interactions to form complex and variable behaviors. However, this theory focuses on the role of the brain or algorithm in intelligence and ignores the role of the physical vehicle.
\par
In 1999, Rolf Pfeifer and Christian Scheier proposed that intelligence arises from the structure and function of an agent’s body, emphasizing the crucial role of the physical body in interaction\cite{2001Pfeifer}. They argued that intelligent behavior originates not only from cognitive processing but also depends on continuous interaction with the environment. This perspective builds on Rodney Brooks’s concept of environmental interaction and highlights the fundamental role of physical embodiment in intelligence, shaping the foundation of embodied intelligence. Rolf Pfeifer and Christian Scheier pointed out that the mobility of animals and robots does not only come from the instructions of a central control system but is a direct result of the interaction between their body structure and specific environmental factors. The morphology and physical properties of the body, such as shape, mass distribution, friction properties, etc., directly affect the generation of intelligent behaviors. 
\par
In 2005, Linda Smith proposed the "embodiment hypothesis," which explores human cognition through the lens of embodiment in cognitive science\cite{2005Smith}. This hypothesis asserts that human cognitive processes do not solely rely on abstract mental processes but emerge through direct interaction with the physical environment\cite{2005Smith}. She argues that the development of human thinking, perception and a variety of other abilities stems from the individual's continuous interaction with the environment. This perspective emphasizes that the body is not only a tool for perceiving the world, but also a key to shaping cognitive structures, such as the way infants learn the laws of physics by touching and manipulating objects or understanding spatial relationships through the coordination of vision and movement. These behaviors exemplify the active role of the body in the cognitive process. The embodied hypothesis further emphasizes that the role of the environment in cognitive development is crucial, arguing that the structure and characteristics of the environment not only provide sensory input but are also directly involved in the formation of cognitive abilities\cite{black2012embodied}.\ For example, the objects and spatial layout of the physical environment affect an individual's action selection and perceptual development, and thus the establishment of cognitive structures\cite{golledge2018environmental}. This role of environmental involvement makes cognitive development tightly linked to specific environmental conditions. Linda Smith's research expands our understanding of cognitive science by shifting the traditional focus on the internal mechanisms of the brain to focus on the dynamic interactions between the body and the environment. However, technical limitations in the early 2000s constrained the practical realization of these principles. Early embodied AI systems struggled with real-time sensory processing and adaptive learning, as conventional algorithms could not efficiently handle high-dimensional sensorimotor data streams.
\subsection{Deep learning technology enables embodied intelligence}
\par
The paradigm shift occurred in 2012 with the maturation of deep learning architectures. The breakthrough performance of deep convolutional neural networks in ImageNet classification demonstrated for the first time that hierarchical feature extraction could autonomously process raw sensory data — a capability essential for materializing embodied intelligence\cite{manakitsa2024review}. This technological evolution enabled three fundamental advancements: (1) Parallel processing of multimodal sensor streams (visual, tactile, proprioceptive) through spatially invariant convolutional operations\cite{li5128520sensor}
; (2) Emergence of self-supervised affordance learning via gradient-based optimization in high-dimensional action spaces\cite{manuelli2020keypoints}
; (3) Development of physics-informed neural controllers capable of adapting to environmental perturbations within 200ms latency\cite{manuelli2020keypoints}
. These technical milestones transformed theoretical embodiment principles into functional systems, as evidenced by multimodal robots achieving high success rate in unstructured object manipulation and embodied simulators demonstrating cross-domain policy transferability\cite{manuelli2020keypoints}
. By overcoming the "curse of dimensionality" that plagued early symbolic approaches, deep learning provided the mathematical framework to operationalize continuous environment-body-cognition loops — the core tenet of Smith's hypothesis.
\subsection{Large language models enable embodied intelligence}
The epistemological transformation of embodied intelligence accelerated through the integration of large language models (LLMs), which introduced three fundamental reconceptualizations:
\par
1.Symbolic Embodiment: LLMs established bidirectional mappings between linguistic abstraction and sensorimotor patterns, enabling the grounding of semantic concepts in physical affordance spaces.
\par
2.Cognitive Scaffolding: The latent space of foundation models provides probabilistic priors for inferring environmental dynamics, effectively bridging the gap between phenomenological experience and formal reasoning.
\par
3.Metacognitive Regulation: Through attention-based architectures, embodied systems acquired the capacity to dynamically reweight sensory inputs and action policies based on contextual salience.
\par
This paradigm shift is best characterized by the emergence of language-mediated embodiment, where the syntactic structure of LLMs constrains the exploration of action-perception manifolds while preserving the open-endedness required for environmental adaptation. The resultant systems exhibit metaplasticity — their interaction histories recursively reshape both the language model's conceptual embeddings and the embodied controller's policy gradients.
\par
Such architectural innovations realize Smith's vision at computational scale: the environment now actively participates in shaping cognitive representations through dual-channel feedback loops (linguistic-conceptual and sensorimotor-procedural). The dialectical synthesis of symbolic priors and physical instantiation has consequently redefined the ontological status of embodied intelligence, transcending traditional dichotomies between connectionist and embodied cognition frameworks.
\subsection{The Role of EI in Achieving AGI}
At this point, the basic principles of embodied intelligence systems have been formed:
\par
1. The system must adapt dynamically to changes in the environment. This highlights that EAI systems must go beyond predefined logic and dynamically adjust to real-time conditions. For example, in a complex and changing real-world traffic surface, an automated navigation robot can autonomously plan its route based on nearby vehicles and pedestrian traffic, rather than relying only on a predefined path.
\par
2.\ The system must incorporate an evolutionary learning mechanism, enabling it to learn from experience and incrementally optimize its performance, in accordance with AGI's principles of autonomous learning. For example, a cleaning robot continuously accumulates experience about which areas are prone to dirt accumulation during the cleaning process, thus optimizing the cleaning route.
\par
3.\ Environmental influence is fundamental to behavior and cognition. The environment not only provides a stage for action but also actively shapes behavior and cognitive development. All components of the environment may profoundly influence the development of the system and the formulation of implementation strategies. Therefore, systems need to understand and leverage the influence of the environment to continuously optimize the most appropriate strategy for the environment, significantly improving adaptability and intelligence. For example, a fire rescue robot can interactively adjust its rescue strategy based on terrain and obstacles to ensure safe and efficient operation.
\par
Evolutionary and learning mechanisms in embodied intelligence systems enable the potential development of intelligence surpassing human capabilities, although this remains a distant objective. The role of the environment in the construction of cognitive behavior is one of the key factors in achieving intelligence. Through continuous interaction with the environment, the system continuously adjusts and optimizes its behavioral strategies, thus gradually improving its intelligence level. This dynamic adaptive learning process endows the system with a high degree of flexibility and adaptability in complex environments, signaling the potential for achieving AGI in the future.

\begin{figure*}[!h]
\centering
\includegraphics[width=0.9\linewidth]{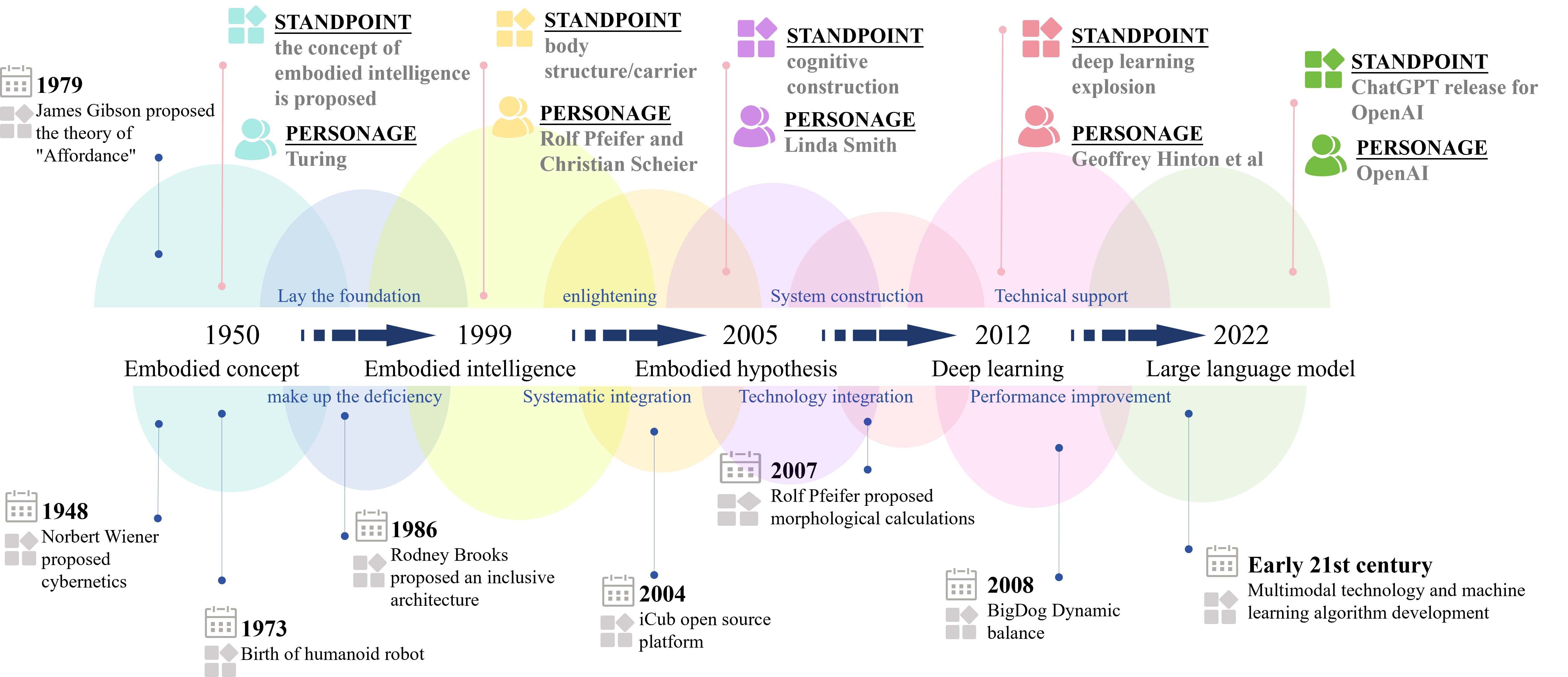}
\caption{The concept of embodied intelligence has gone through three important stages of development, culminating in today's definition; the concept embodies three fundamental principles for system-level architectural design. Although the concept was once marginalized, with breakthroughs in robotics, reinforcement learning, and multimodal learning, embodied intelligence has regained widespread attention in the context of the rapid development of artificial intelligence.}\label{Fig1}
\end{figure*}

\section{Technological Frameworks of Embodied Intelligence}
\par
The implementation of embodied intelligence systems has coalesced around two primary architectural paradigms: end-to-end frameworks and modular decomposition frameworks. These approaches diverge fundamentally in their design philosophy, implementation complexity, and adaptability to real-world scenarios, while sharing the common goal of enabling intelligent physical interactions.The differences between the two are shown in Table~\ref{tab2}.

\begin{table}[H]
\fontfamily{ptm}\selectfont
\centering
\caption{Detailed differences between Modular Architecture and End-to-End Architecture}
\begin{tabular}{ p{1.5cm} p{2.6cm} p{2.6cm}}\toprule
 \hline
  Dimension & Modular Architecture  & End-to-End Architecture   \\
 \hline
  Development Cost & Low (independent module optimization, suitable for small teams) & Extremely high (requires billions of data samples and thousands of GPUs) \\
  \hline
 Performance Ceiling & Limited by module interdependency ("Bucket Effect") & Theoretically higher (global optimization eliminates human design bias)\\ 
 \hline
 Interpretability & High (clear module functionality) & Low (black-box model relying on emergent capabilities) \\
 \hline
 Main Adopters & Academia, startups (research teams) & Tech giants (OpenAI, Google, Tesla) \\
 \hline
  Typical Applications & Industrial automation, task-specific robots & General-purpose robots, complex environment interaction \\
\hline
\end{tabular}
\label{tab2}
\end{table}

\subsection{Framework Definitions and Core Characteristics}
\par
The end-to-end framework operates through a unified neural architecture that directly maps raw sensory inputs (e.g., vision, LiDAR, proprioception) to actuator control signals. This paradigm eliminates explicit functional boundaries between perception, decision-making, and action generation, relying instead on deep neural networks to implicitly learn intermediate representations. In contrast, the modular decomposition framework adopts a hierarchical structure, systematically dividing the system into four discrete components: 1.Perception Module(Environmental state reconstruction through sensor fusion) 2.(Decision-Making Module: Task planning and strategy optimization) 3.Action Module:(Kinematic/dynamic motion generation) 4.Feedback Module(Performance evaluation and online adaptation)
\par
The critical distinction lies in their information processing mechanisms: end-to-end systems employ global optimization across the entire sensorimotor chain, while modular systems achieve localized optimization through predefined interface specifications.
\subsection{End-to-End Framework: Current Progress and Industrial Implementation}
\par
The advancement of end-to-end embodied intelligence frameworks (e.g., Vision-Language Models/VLMs and Vision-Language-Action/VLA models) currently faces a critical bottleneck: data acquisition and scalability. Three primary data generation paradigms have emerged across the industry: 1.Physical-world data collection. 2.Virtual simulation synthesis. 3.Hybrid physical-virtual data fusion.

\subsubsection{Architectural Innovations}
\par
Tesla's FSD V12: exemplifies pure physical-world learning, utilizing temporal convolutional networks trained on multi-modal sensor streams from millions of vehicles. This paradigm bypasses explicit path planning through emergent spatial-temporal reasoning learned directly from real-world driving behaviors.

\par
{Huawei's GOD-PDP Architecture: adopts a hybrid approach, combining real-world obstacle detection (General Obstacle Detection) with simulated decision processors (Predictive Decision Processor). Its "neural subgraph" design enables gradual migration from rule-based to data-driven control while maintaining operational safety.

\par
XPeng's XPlanner: demonstrates China's distinctive strategy by establishing dedicated humanoid robot training centers in Beijing. These facilities collect teleoperation data through synchronized motion capture systems, generating annotated demonstration trajectories for urban navigation tasks.

\subsubsection{Technical Challenges and Data Solutions}
\par
Current implementations address the data scarcity challenge through geographically differentiated approaches:
\par
U.S. Synthetic Focus: Waymo's neural scene diffusion models generate photorealistic collision scenarios, augmenting real-world data by 40× through parametric accident reconstruction.
China's Teleoperation Infrastructure: Companies like UBTECH deploy AR glass-assisted data collection systems, capturing 6D pose annotations and force feedback signals during human demonstration tasks.
Cross-domain Adaptation: Baidu Apollo bridges the simulation-to-reality gap by joint training on driving videos and traffic regulation texts, aligning latent representations between virtual and physical domains.

\subsubsection{Industrial Adoption Trends}
\par
Regional implementation patterns reveal fundamental philosophical divergences:
\par
U.S. Simulation Dominance: Leveraging established simulation ecosystems (CARLA, NVIDIA DRIVE Sim), American firms prioritize virtual data synthesis for rare-edge case coverage.
China's Physical-first Strategy: National humanoid robot innovation centers employ fleet-scale teleoperation, combining visual demonstration parsing with proprioceptive data recording.
Hybrid Convergence: European OEMs like Volkswagen deploy "digital twin" pipelines that mirror Chinese teleoperation data into synthetic environments, achieving 78\% cross-domain policy transferability.
\section{Four Core Components of Embodied Intelligence }
The definition of AGI remains controversial, leading to variability in defining its objectives. This paper adopts the six AGI principles proposed by DeepMind in 2023 as criteria for evaluating the core modules of embodied intelligence\cite{2023Morris}. Based on recent research and authoritative sources, we analyze how each module contributes to implementing these principles. The six principles include: focusing on model capabilities, not processes; focusing on generalizability and performance; focusing on cognitive and metacognitive tasks; focusing on potential, not deployment; focusing on ecological validity; and focusing on the AGI development path, not just the end point. A detailed explanation can be found in the article Levels of AGI: Operationalizing Progress on the Path to AGI. The discussion in this chapter will build on these six principles to illustrate how embodied intelligence modules can put them into practice.
\par
Sun Fuchun, a professor at Tsinghua University, pointed out at the 2024 Digital Intelligence Technology Conference that EAI is generally considered to be composed of four core elements: ontology, intelligence, data and knowledge, learning and evolutionary structure. As shown in Figure~\ref{Fig1}, ontology acts as a multimodal perception-action interface, providing an embodied interaction channel of the physical world. Based on the closed-loop of perception, reasoning and decision, the agent constructs the environment cognitive model to drive the generation of adaptive behavior. The virtual-real fusion data ecology and the domain knowledge graph together form the experience base of the system, and the data barriers are broken through the migration mechanism. The hierarchical reinforcement learning and meta-evolution architecture form a dynamic optimization engine, which enables the system to realize the autonomous iteration of cognitive strategies and ability generalization in continuous environment interaction, and finally achieve intelligent emergence under physical constraints. 
\par

\begin{figure*}
\centering
\includegraphics[width=0.9\linewidth]{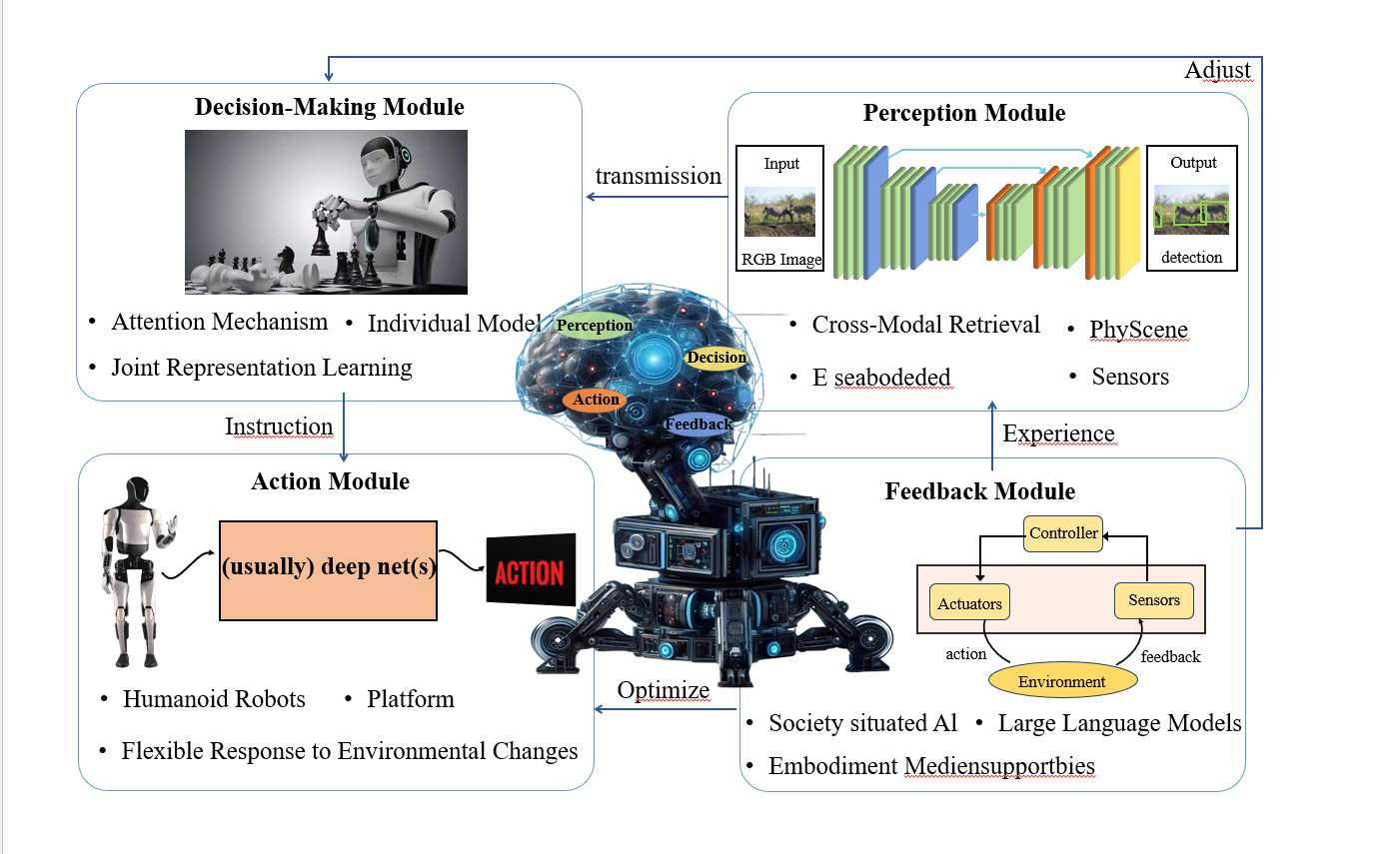}
\caption{An intelligent agent is composed of four major modules: the decision-making module, the perception module, the action module, and the feedback module\cite{2021Hegde}\cite{2014Nogueira}\cite{2020Xia}\cite{2021Coumans}\cite{2022Fernandez-Chaves}\cite{2007Ganapathy}\cite{coumans2021pybullet}. The feedback module can conduct real-time monitoring and accumulate experience for the perception module, enabling the perception module to collect real-time environmental information, transmit the information to the decision-making module, and simultaneously adjust the operation of the decision-making module so that it can generate decisions for actions in the environment. The action module\cite{2018Shigemi} receives instructions from the decision-making module and optimizations from the feedback module, and controls the physical carrier to complete various actions.}\label{Fig2}
\end{figure*}

According to the cooperation of the four elements in the whole embodied intelligence system, the workflow of embodied intelligence can be divided into perception module, decision-making module, action module, and feedback module. These modules form a highly integrated closed-loop architecture that supports the sophisticated flow of information and actions among the modules. As shown in Figure~\ref{Fig2}, The perception module collects real-time environmental data from its physical components and transmits it to the intelligent decision-making module. This module integrates information from the perception system, simulation environments, open-source data, and other relevant sources to generate an action strategy, which is then relayed to the action module. The action module accurately executes the task according to the decision instructions, prompts the physical carrier to complete the specific operation, and provides feedback on the progress of the task in the process. The feedback module monitors the task execution effect in real time and transmits the feedback information back to the decision module to support the system to adjust and optimize the decision strategy.
\par
In this chapter, we will delve into the four core modules of embodied intelligence systems - perception module, decision module, action module, and feedback module - and analyze how their collaborative approach can contribute to the development of AGI. The corresponding taxonomy tree is shown in Figure~\ref{Fig3}. We adopt different technological categorizations with the aim of providing a diversified perspective on the field. Through the collaboration of the four modules, the embodied intelligence system can not only enhance the adaptability and flexibility of the intelligences, but also improve their multitasking ability in complex environments, laying the foundation for the realization of AGI. The role of each module in realizing DeepMind's six principles will provide theoretical support and technical paths for the application and development of embodied intelligence in the field of AGI.

\subsection{perception module}
The physical interaction between the entity and the environment forms the foundation of embodied perception\cite{2020Liu}. Embodied perception primarily evolves into visual recognition, auditory perception, and tactile perception, among others. According to research, various perceptual modalities are currently trending towards multimodal fusion, which is a crucial prerequisite for achieving embodied intelligence. Multimodal fusion compensates for the limitations of single perceptual modalities by integrating multiple sensory information, thereby enhancing the system's understanding and adaptability to the environment. For instance, visual perception provides spatial information, while tactile perception supplements the physical characteristics of object surfaces; the combination of both can lead to more accurate identification and manipulation of objects. This paper will not delve into the details of visual, auditory, and tactile perception but will instead focus on conducting a detailed survey of the significant technology of multimodal perception.
\par
Given the complexity of human-like multimodal perception, coordinating perceptual modules such as vision, touch, and hearing remains a major challenge in robotic technology development. The introduction of multimodal models offers a novel approach to addressing this issue by integrating data from different sensory channels, thereby supporting information analysis and in-depth understanding \cite{2024Qin}, and serving as a core technical pillar for embodied intelligent systems. This collaboration not only significantly enhances the perceptual accuracy of robots but also provides greater adaptability and robustness for task execution in complex environments. As technology continues to evolve, the processing capabilities of intelligent systems in various complex scenarios are constantly improving, approaching the universal adaptability principle of AGI (Artificial General Intelligence). Moreover, cross-modal collaboration and information fusion enhance the transparency and interpretability of systems, enabling robots to provide more intuitive feedback and improving the comprehensibility and trustworthiness of human-robot collaboration. The aforementioned research indicates a growing trend towards multimodal fusion in perceptual modules, laying an important foundation for the content of subsequent chapters.
\par
The essence of multimodal technology lies in the efficient integration of data from different modalities, overcoming their disparities and heterogeneity, and fully leveraging the strengths of each modality to achieve performance superior to that of any single modality\cite{2021Wang}. The perceptual process of multimodal models encompasses six critical steps: data acquisition, preprocessing, feature extraction and representation, data fusion, multimodal learning and inference, and output and decision-making (as illustrated in Figure~\ref{Fig3}). In the following sections, each of these steps will be elaborated upon with specific examples.

\begin{figure*}[htbp]
\centering
\includegraphics[width=0.7\linewidth]{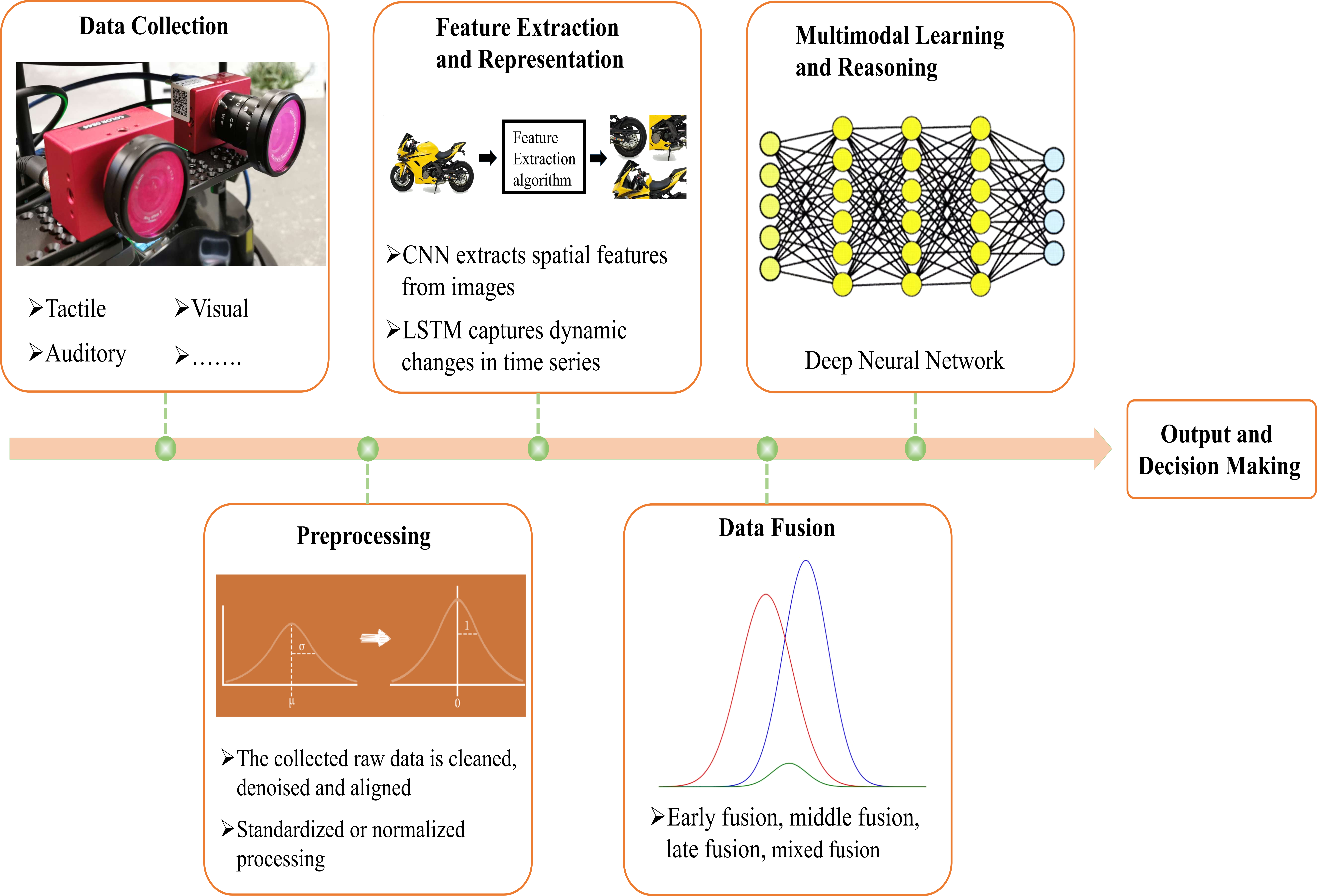}
\caption{The perception process of multimodal models typically involves five steps: The first step is data acquisition  \cite{2014Winkler}\cite{2017Zou}\cite{2018Guo}, which aims to collect multidimensional perceptual data through various sensors; the second step is data preprocessing \cite{2019Shorten}\cite{2002Elaksher}\cite{2016García}, where the collected raw data undergoes cleaning, noise reduction, alignment, and other processes to ensure consistency across modalities; the third step is feature extraction and representation \cite{2017Wu}\cite{2021Lindemann}\cite{2014Luo}, where effective features are extracted from the raw data of each modality; the fourth step is data fusion, where data from different modalities are merged to enhance overall perception; and the fifth step is multimodal learning and inference, where the fused features are used for higher-level tasks. }\label{Fig3}
\end{figure*}

1)Data Acquisition: data acquisition is the first step in multimodal sensing, aiming to acquire data in different modalities through multiple sensors \cite{2022Huang}, which usually originate from sensors in different modalities, such as visual sensors (cameras), haptic sensors (force sensors, flex sensors) and auditory sensors (microphone arrays). The sensors differ in format, sampling frequency, dimension and accuracy. For example, in autonomous driving, image data provided by cameras differ significantly from LiDAR (laser radar) data in terms of spatial resolution and scale.
\par
2)Data preprocessing: the preprocessing stage mainly deals with cleaning, noise reduction and temporal alignment of the collected raw data to ensure the consistency of the modal data \cite{2022Younis}. For example, visual and haptic data need to be temporally aligned to ensure that their state data can correspond at the same point in time. In addition, to address the differences in scale and units of different modal data, they need to be normalized or normalized. In multimodal autonomous driving systems, image data and LiDAR data need to be spatially aligned to accurately reflect the same scene.ata need to be spatially aligned to accurately reflect the same scene.
\par
3)Feature extraction and representation: feature extraction is one of the key steps in multimodal perception \cite{2024Almujally}, where key features are extracted from raw data of different modalities by deep learning models (e.g., convolutional neural network CNN, recurrent neural network RNN, long-short-term memory network LSTM, etc.).\ CNNs are suitable for extracting spatial features such as edges, textures, and colors from visual data; LSTMs are suitable for mining tactile time-series dynamic features in signals. For example, in an object detection task for autonomous driving, CNN extracts object features in the image, while LSTM is used to understand the object's motion trajectory.
\par
4)Data fusion: data fusion is one of the core steps in multimodal perception, which improves the overall perception by merging data from different modalities \cite{2015Simanek}. Common fusion methods include early fusion, intermediate fusion, late fusion and hybrid fusion. Early fusion directly merges the original data in order to utilize the complementary information to enhance the generalization ability\cite{2020Bednarek} , for example, in the task of sentiment analysis, visual, speech and text data are often directly combined for training; mid-term fusion merges the features of each modal data after feature extraction; in recent years, mid-term fusion based on the self-attention mechanism and graph neural networks have been widely used\cite{2023Lin} , which can dynamically assign It can dynamically assign different weights to different modal features and effectively capture the non-Euclidean relationship between modalities, and is commonly used with the self-attention mechanism and graph neural networks; Late fusion combines the data at the decision-making level after the modal data are processed independently, which is suitable for the case of large modal differences.Recent research advances have seen Generative Adversarial Networks (GANs) being used to simulate inter-modal feature generation \cite{2020Roheda} and reinforcement learning optimizing fusion strategies. These methods have played an important role in improving the accuracy, robustness and adaptability of multimodal perception \cite{2018Nadon}.
\par
5)Multimodal Learning and Reasoning: in this stage, the fused features are used for higher level tasks such as object recognition, action recognition, semantic understanding, etc \cite{2009Noceti}. Techniques such as deep neural networks (DNN) and graph neural networks (GNN) are widely used in this stage \cite{2024Jia}. For example, in video understanding, CNNs and LSTMs extract visual and time-series features, while GNNs are capable of in-depth reasoning about relationships between objects. In autonomous driving, multimodal learning combines camera images and radar data to improve object detection and path planning accuracy \cite{2022de}.
\par
Based on the multimodal learning results, the perception module passes the obtained tasks and instructions to the decision-making module for further refinement and operation \cite{2003Fritsch}. In the face of complex multidimensional perceptual environments, how to coordinate visual, tactile, auditory, and other multisensory modules and realize collaboration has become an important challenge in the development of current technologies\cite{2023Tang}. The introduction of multimodal modeling provides a possibility to solve this problem. The multimodal model realizes comprehensive information analysis and deep understanding by fusing data from different sensory channels. With the help of multimodal technology, the embodied intelligence system can significantly improve the adaptivity and multitasking ability through the cooperative work with other modules, which aligns closely with the ultimate goal of AGI. From the perspective of the six principles of AGI, the multimodal perception module enhances the system's environmental understanding by integrating multiple sensory data and makes the model applicable to a variety of tasks to enhance versatility. In addition, the system focuses on learning potential and continuously improves performance through data accumulation and model optimization.

\subsection{Decision-making module}
The decision-making module receives information from the perception module and interaction data from the feedback module, performs task planning and decision generation, and transmits the decision results to the action module for execution \cite{2018Schwarting}. In recent years, the rapid development of deep learning and reinforcement learning has significantly advanced the capabilities of the decision-making module \cite{2023Singh}, enabling it to play a pivotal role in the path toward Artificial General Intelligence (AGI). The decision-making module consists of four primary functions: environmental understanding and reasoning, task planning, decision generation, and a learning and evolution framework \cite{2016Tsarouchi}\cite{2000Dietterich} (as shown in Figure~\ref{Fig4})).

\begin{figure}[htbp]
\centering
\includegraphics[width=0.9\linewidth]{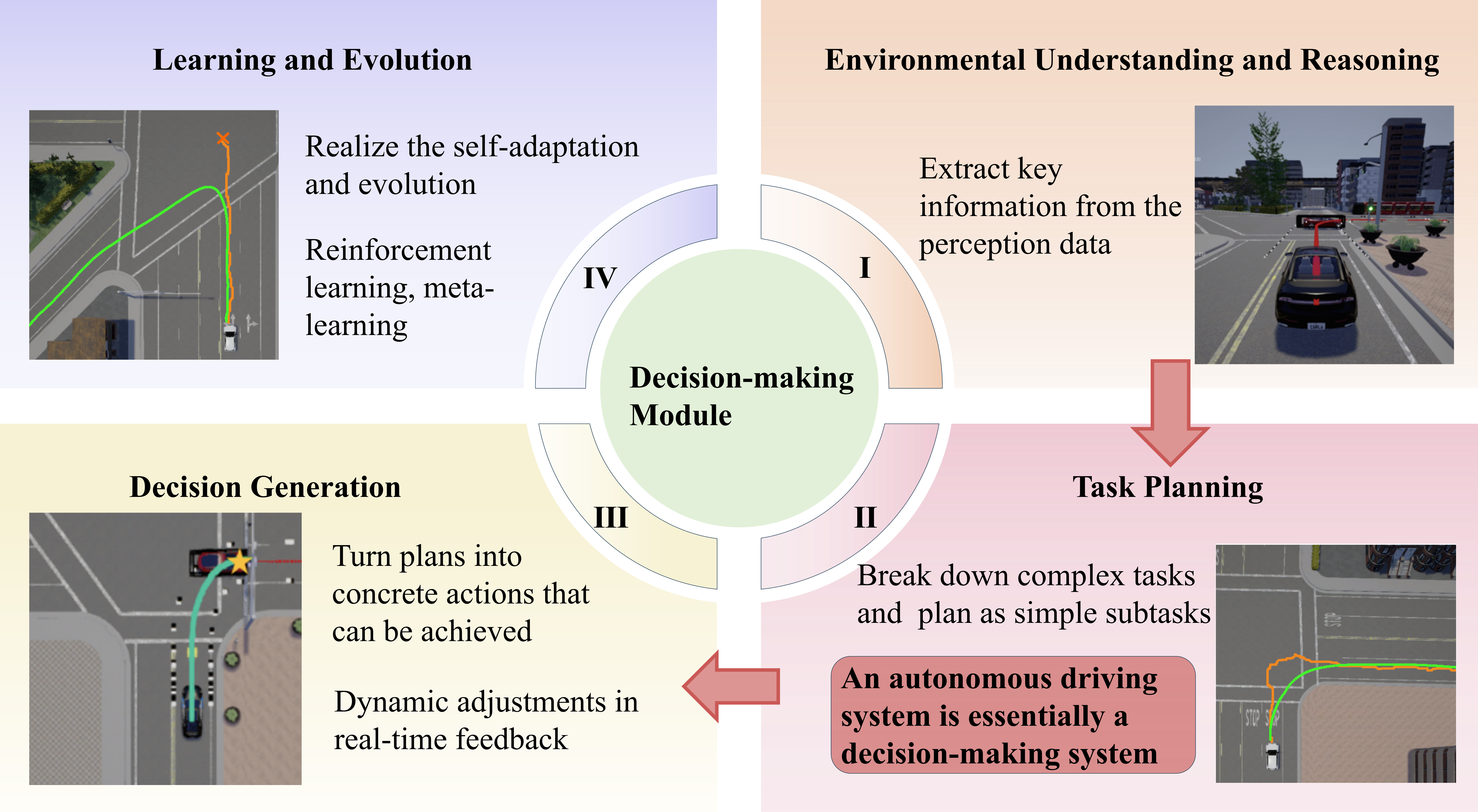}
\caption{The decision-making module comprises four main functional components: environmental understanding and reasoning, task planning, decision generation, and learning and evolution. The environmental understanding and reasoning module \cite{2004Michel}\cite{2006Hohl} extracts critical information from perceptual data to construct a comprehensive understanding of the current environment. The task planning module \cite{2008Galindo}\cite{2006Alami} is responsible for decomposing tasks and formulating plans. The decision generation module \cite{2019Paxton} selects and executes specific actions based on the plans. These three components work closely together to ensure the efficiency and accuracy of the decision-making process. The learning and evolution framework \cite{2021Gupta}\cite{2004Floreano} permeates the entire decision-making module, continuously optimizing decisions through methods such as reinforcement learning and meta-learning, leveraging environmental feedback to promote the adaptation and evolution of the intelligent agent\cite{huang2021learning}.}\label{Fig4}
\end{figure}
\par
First, the environmental understanding and reasoning module, as the starting point of the decision-making process, is primarily tasked with extracting key information from perceptual data and constructing a comprehensive understanding of the current environment \cite{2021Sadhu}. This module embodies the AGI principle of "focusing on generality and performance," as it is capable of generating efficient planning strategies across various task scenarios. The task planning module is responsible for decomposing complex tasks into a series of subtasks and formulating rational plans \cite{2023Guo}. The decision generation module selects and executes actions based on the planning results, aiming to translate plans into executable actions and dynamically adjust them in response to real-time feedback \cite{1999Sutton}. These three modules are sequentially interdependent in the workflow, ensuring that the decision-making process is both efficient and accurate \cite{2016Ho}.
\par
The learning and evolution framework permeates the entire decision-making module, leveraging methods such as reinforcement learning and meta-learning to continuously optimize the decision-making process through environmental feedback \cite{2021Hospedales}\cite{2017Salimans}. This enables the decision system to adapt to new challenges and task variations, thereby achieving the self-adaptation and evolution of the intelligent agent. This mechanism aligns with the AGI principle of "focusing on potential rather than deployment," meaning the system enhances its capabilities through continuous learning rather than relying solely on predefined rules. Additionally, the self-learning and evolutionary characteristics of the decision-making module reflect the long-term goal-oriented principle, ensuring that the intelligent agent not only handles current tasks but also accumulates experience, optimizes future decisions, and adapts to complex and dynamic environments.
\par
Next, the latest advancements and future development trends of the four functional modules will be introduced. The most recent progress is summarized in Table~\ref{tab3}.

\begin{table}[!h]
    \centering
   \begin{tabular}%
       {| >{\centering\arraybackslash}p{0.15\textwidth }
       | >{\centering\arraybackslash}p{0.05\textwidth }
       | >{\centering\arraybackslash}p{0.2\textwidth }|}
         \hline
        Functional Module & Year & Innovative Industries\\
        \hline
        \multirow{7}{0.1\textwidth}{Environmental understanding and reasoning} & \makecell[c] {2023} & \makecell[c] {TVR\cite{2023Hong}} \\ 
             \cline{2-3}
             ~ & 2020 & deep convolutio nal recurrent neural network\cite{2020Driess} \\
             \cline{2-3}
             ~ & 2024 & Perceptual Fusion\cite{20Li24} \\
             \cline{2-3}
             ~ & 2024 & DAGNN\cite{2024Cai} \\
             \cline{2-3}
             ~ & 2024 &  T W o-phase K AL ma n Filter with U ncertainty quan T ification\cite{20Zhou24} \\
             \cline{2-3}
             ~ & 2022 & HSSN\cite{2022Li} \\
             \cline{2-3}
             ~ & 2023 & DeepAdaIn-Net\cite{2023Wang} \\
             \cline{2-3}
         \hline
         \multirow{5}{0.1\textwidth}{Mission planning} &2023 & GRID\cite{2023Okubo} \\ 
            \cline{2-3}
            ~& 2024 & multi-USV task planning method based on improved DRL\cite{2024Zhang} \\
            \cline{2-3}
            ~ & 2024 & SkillDiffuser\cite{2024Liang} \\
            \cline{2-3}
            ~ & 2024 & MAPF\cite{2024Klar} \\
            \cline{2-3}
            ~ & 2024 & combines adversarial imitation learning with LLM\cite{2024Sun} \\
         \hline
          \multirow{6}{0.1\textwidth}{Decision Generation} &2024 & DCEA-DQN\cite{20Liang24} \\ 
            \cline{2-3}
            ~ & 2024 & FSDD-MAML\cite{20Liu24Liu} \\
            \cline{2-3}
            ~ & 2024 & Hybrid Inverse Reinforcement Learning\cite{2024Ren} \\
            \cline{2-3}
            ~ & 2024 & framework predicated on dynamic and socially-aware decision-making game theory\cite{2024LiuJ} \\
            \cline{2-3}
            ~ & 2024 & Adaptive Algorithms\cite{2024Sahoo} \\
            \cline{2-3}
            ~ & 2024 & MBRL\cite{2024Dong} \\
          \hline 
          \multirow{5}{0.1\textwidth}{A Framework for Learning and Evolution} &2024 & Deep Reinforcement Learning\cite{2024Shuford} \\
            \cline{2-3}
            ~ & 2024 & MAMBA\cite{2024Rimon} \\
            \cline{2-3}
            ~ & 2019 & Multi-agent Adversarial Reinforcement Learning\cite{2019Wachi} \\
            \cline{2-3}
            ~ & 2019 & Large Scale Incremental Learning\cite{2019Wu} \\
            \cline{2-3}
            ~ & 2023 & Forward-Forward Algorithm for Self-Supervised Learning\cite{2023Brenig} \\
        \hline
        \end{tabular}
     \caption{Summary of the latest progress of the decision module}
    \label{tab3}
\end{table}
\subsubsection{Environmental understanding and reasoning}
The core task of the environmental understanding module is to extract key information from perceptual data and construct a comprehensive understanding of the current environment. This primarily involves the processing and interpretation of perceptual data as well as environmental modeling.
\par
Perceptual data processing and interpretation involve extracting meaningful information from raw data collected by various sensors (such as cameras, LiDAR, sonar, inertial measurement units, etc.) \cite{2019Wang}. This process typically includes steps such as data preprocessing, feature extraction, data fusion, and information interpretation. During the data preprocessing stage, tasks like noise reduction, filtering, and normalization are often performed to ensure data quality and accuracy. Feature extraction employs deep learning or traditional computer vision algorithms (e.g., CNN) to identify key features from the data, such as object boundaries, textures, and shapes. Data fusion is used to integrate data from different sensors, addressing the limitations of single-sensor environmental perception \cite{2016Cadena}. For example, the fusion of visual and LiDAR data enhances the accuracy of perceiving object positions and motion states. Finally, information interpretation utilizes algorithms such as deep neural networks and reinforcement learning to analyze patterns and relationships within the perceptual data, enabling tasks like object recognition, scene understanding, and behavior prediction.
\par
		The environmental modeling aspect focuses on how to model the environment (e.g., 3D environment model, state estimation, etc.) by processing sensor data, which mainly includes tasks such as 3D environment model construction, state estimation and map generation. Traditional environmental modeling techniques, such as image-based feature extraction and 2D map generation, usually rely on simpler algorithms, such as SIFT (Scale Invariant Feature Transform) and SLAM (Simultaneous Localization and Map Construction) techniques. These methods are able to extract basic information about the environment from sensor data and construct simple maps based on sensor feedback. However, the limitations of traditional methods are their poor accuracy and robustness, and they usually cannot effectively handle 3D spatial information. With the advancement of deep learning and multimodal sensing technologies, new 3D environment modeling and state estimation methods are emerging. For example, deep learning-based 3D reconstruction techniques are able to automatically recover a more accurate 3D environment model from image data from multiple viewpoints through models such as deep convolutional neural networks (CNN) and generative adversarial networks (GAN), which can effectively overcome the effects of illumination and viewpoint changes and improve the accuracy and stability of 3D reconstruction.\ Meanwhile, state estimation methods based on deep learning, such as vision-based SLAM (V-SLAM) and deep image processing, can more accurately estimate the state of a robot or an intelligent body, including the position, velocity, and attitude by fusing data from different sensors (e.g., camera, LIDAR, IMU, etc.). For map generation, the traditional LIDAR-SLAM (LIDAR-SLAM) technology has shown strong capability in constructing 2D or 3D maps that can accurately capture static and dynamic obstacles in the environment.\ However, the adaptability of these techniques in complex and dynamic environments is still limited, especially in rapidly changing scenes that are prone to errors. DeepSLAM (DeepSLAM) is able to model the environment more accurately and respond effectively to dynamic changes by automatically extracting and optimizing environmental features through deep neural networks. In addition, multimodal SLAM technology further improves the accuracy of map generation by fusing LiDAR, camera, and other sensor data, especially in structurally irregular or dynamic environments, allowing for more flexible and adaptive adjustments.

\subsubsection{Mission planning}

Task planning is mainly responsible for transforming high-level task goals into executable subtasks and formulating optimal execution paths. Task planning not only needs to take into account the dynamic changes of the environment, but also coordinates the collaboration among multiple intelligences. In the past, traditional task planning methods mostly relied on rule and model reasoning to decompose and schedule tasks through manually designed algorithms. However, these traditional methods are usually limited by task size, dynamic environment changes, and computational resources, making it difficult to meet the demands of complex and large-scale tasks. With the introduction of techniques such as deep learning and reinforcement learning, the effectiveness and adaptability of task planning have been significantly improved.
\par
	Traditional task planning techniques usually include algorithms based on graph search (e.g., A*, Dijkstra) and constraint planning methods. These methods solve the task decomposition and path planning problems by defining explicit goals and paths for each task. However, these techniques typically exhibit poor robustness in the face of dynamically changing environments. For example, traditional graph search methods need to recalculate paths when the environment changes (e.g., when obstacles appear or environmental conditions change), while rule-based planning methods often become computationally complex and inefficient when the task complexity increases.In addition, these techniques often struggle with coordinating and enabling cooperation among multiple intelligent agents, especially in complex and uncertain environments with poor efficiency and stability.
\par
    In recent years, cutting-edge technologies such as deep learning and reinforcement learning have been gradually introduced into the task planning field, greatly expanding the application scope and capability of traditional methods.\ GRID (Graph-based Reinforcement Learning for Intelligent Decision-making) is a graph-based reinforcement learning approach \cite{IEEEMulti-agent}, which aims to train intelligences to learn efficient decision-making strategies in complex environments through reinforcement learning.\ Compared with traditional methods, GRID can automatically adapt to changes in the environment and provide more flexible and efficient mission planning. The improved Deep Reinforcement Learning (DRL)-based multi-Unmanned Vehicle (multi-USV) mission planning approach combines Deep Reinforcement Learning (DRL) with multi-intelligent body collaboration for multi-USV mission planning \cite{2023Multi-usv}. In this framework, multiple intelligences coordinate their respective tasks and paths through joint learning, and are able to dynamically adjust their strategies to cope with complex environmental changes.\ This approach can significantly improve the flexibility and adaptive capability of task planning compared to traditional methods.\ SkillDiffuser is an emerging technique for task planning through diffusion modeling, which particularly excels in handling tasks with long-term dependencies and complex behaviors. By simulating the behavior of intelligences in various environments, this method is able to learn efficient task planning strategies without explicit annotation and generate a series of adaptive task execution plans.\ Multi-intelligent body path planning (MAPF) techniques ensure that the tasks of multiple intelligences in shared environments can be accomplished efficiently and safely by coordinating their movements. In recent years, MAPF methods have significantly improved the planning efficiency of multi-intelligent body systems in complex scenarios, especially in dynamic environments and in the presence of a large number of obstacles, by introducing heuristic search and reinforcement learning algorithms. The task planning approach combining Adversarial Imitation Learning and Large Language Model (LLM) combines Adversarial Imitation Learning (AIL) with Large Language Model (LLM), which is utilized to generate strategies and actions and further optimize the behavior of the intelligences through imitation learning.In the dynamic environment of complex tasks, this approach allows for more flexibility in generating and adapting mission planning scenarios and effectively handling long-term dependencies.
\par
    The introduction of these emerging technologies not only promotes the advancement of task planning technology, but also provides an important path for embodied intelligent systems to realize AGI. Through the combination of deep reinforcement learning, multi-intelligent body collaboration, skill diffusion, and large language modeling, the task planning module is able to exhibit higher adaptivity and flexibility in complex dynamic environments, gradually approaching the ultimate goal of AGI.

\subsubsection{Decision Generation}

Decision generation is the core process by which intelligences generate and execute specific behaviors based on the results of task planning. Its goal is to transform high-level task goals into specific executable actions and make dynamic adjustments based on real-time feedback. With the continuous development of deep learning, reinforcement learning and other technologies, the research on decision generation has gradually shifted from traditional rule-based decision-making methods to adaptive, collaborative and multimodal dynamic decision-making systems. This evolution not only improves the task execution capability of intelligences, but also provides important support for the realization of AGI, especially in the aspects of “focusing on capability, not process” and “focusing on generality and performance”, which show significant advantages.
\par
	In recent years, innovative approaches to decision modules have made significant progress in task planning and decision generation for robots \cite{2022Cooperative}. From the initial rule-based task generation methods to the technological breakthroughs of introducing Deep Reinforcement Learning (DRL) and adaptive algorithms, the capabilities of decision modules have been gradually enhanced. In the early days, Deep Q Learning (DQN)-based decision generation methods provided a more basic decision-making framework for task planning, and with the development of the technology, DCEA-DQN optimized the policy generation in the decision-making process through the introduction of evolutionary algorithms, which further enhanced the adaptability in dynamic environments \cite{PMLRFan}.
    \par
	At the same time, reinforcement learning-based meta-learning methods (FSDD-MAML) have gradually become an important part of decision generation, especially in the context of facing rapid changes and lack of labeled data, by introducing less-sample learning, which enables intelligences to adapt quickly in new tasks and environments, and improves the flexibility and accuracy of decision-making \cite{PMLRZintgraf}. In complex environments with multi-intelligent collaboration and high dynamics, the Game Theory-based Decision-making framework for dynamic and socially aware decision-making introduces a multi-intelligent interaction model for decision generation, which enables more efficient collective decision-making by optimizing the collaboration and competition strategies among individuals.
	\par
    Further progress is embodied in Hybrid Inverse Reinforcement Learning, which solves the problem of scarcity of reward signals in task planning and optimizes strategy generation for complex tasks by learning from expert behaviors and combining it with inverse reinforcement learning. Driven by these techniques, the decision module is not only able to make accurate decisions in more complex environments, but also realizes adaptive adjustment between tasks, which greatly improves the task execution ability of intelligences in complex and dynamic environments. This ability to learn from expert behavior is highly consistent with the principle of “focusing on cognitive and metacognitive tasks”, which enables intelligences to generate decisions more in line with human expectations through imitation and reasoning.
	\par
    As the capabilities of decision modules continue to be enhanced through these innovative approaches, their integration and application in complex scenarios have become increasingly sophisticated. These innovative approaches gradually integrate multiple advanced technologies such as deep learning, game theory, reinforcement learning, etc., enabling decision generation to evolve from simple single-task execution to complex decision systems that can handle multi-tasks and multi-intelligentsia interactions. This evolution not only promotes the development of embodied intelligent systems, but also provides an important path for realizing AGI. Through adaptive, collaborative, and multimodal decision generation technologies, intelligences are able to exhibit higher flexibility and versatility in complex dynamic environments, gradually approaching the ultimate goal of AGI. Such technological development not only embodies the principle of “focusing on the path to AGI rather than a single endpoint”, but also lays a solid foundation for a wide range of real-world applications of intelligences.

\subsubsection{ A Framework for Learning and Evolution}
The learning and evolution framework is the core mechanism for autonomous learning and evolution of intelligences in complex environments. Its main goal is to continuously optimize the decision-making strategy through the continuous interaction between the intelligent body and the environment to cope with the changing task demands and environmental conditions, as shown in the scenario illustrated in Figure~\ref{Fig5}. In recent years, with the emergence of Deep Reinforcement Learning (DRL) and other state-of-the-art methods, learning and evolutionary frameworks have made significant progress in both principle and application. The following techniques are representative within this field, all of which innovate and break with traditional learning methods.
\begin{figure*}[htbp]
\centering
\includegraphics[width=0.7\linewidth]{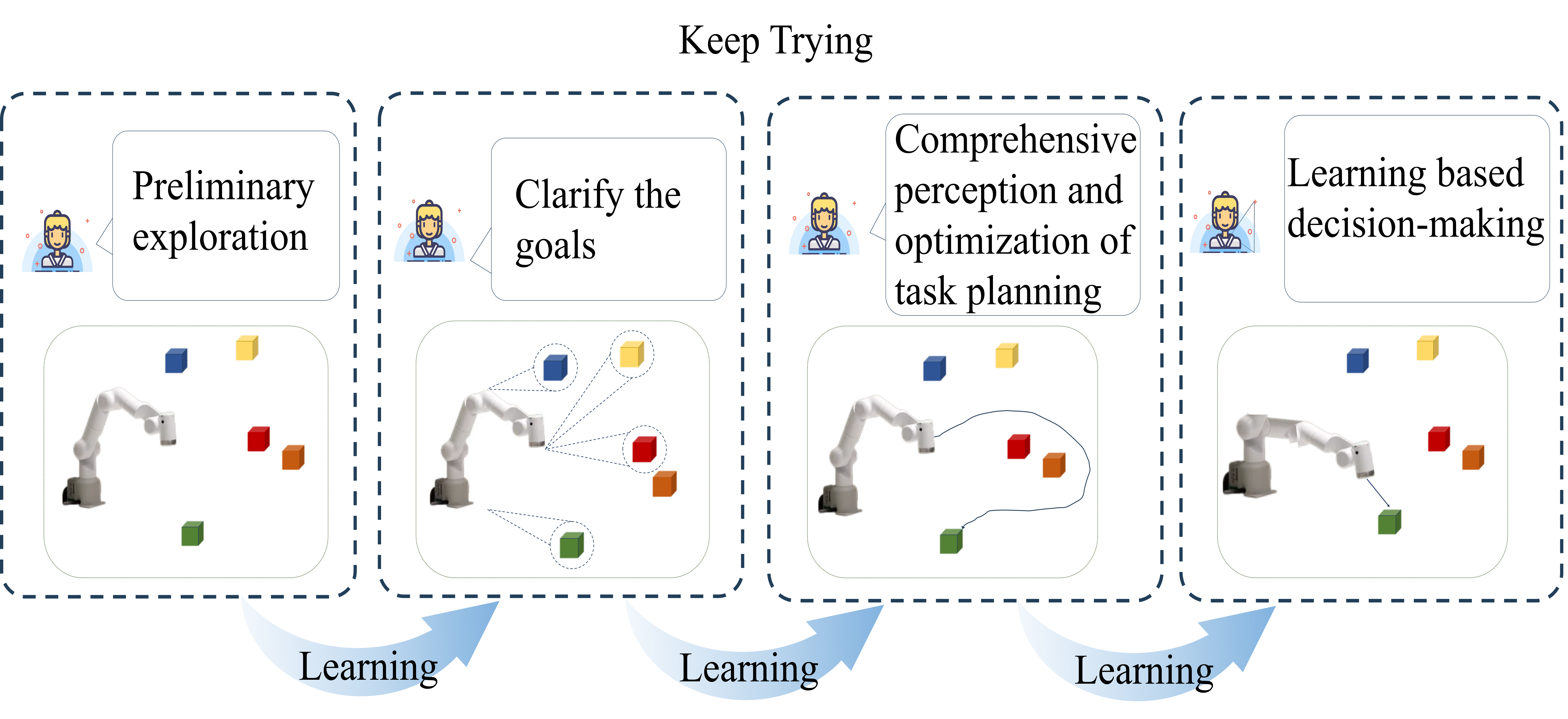}
\caption{The three main functions under the decision module are environment understanding and reasoning, task planning and decision generation. Through step-by-step continuous learning, the sensors complete the perception and comprehensive modeling of the environment, and transform high-level task objectives into executable subtasks, and formulate optimal execution paths, and finally generate and execute specific behaviors.}\label{Fig5}
\end{figure*}

	Deep Reinforcement Learning (DRL) has made rapid development in recent years.The principle of DRL lies in adjusting the behavioral strategies of an intelligent body through environmental feedback, using reward signals to guide the intelligent body to optimize the decision-making process \cite{2015Mnih}. Specifically, the intelligent body uses a deep neural network for policy estimation through interaction with the environment, which is updated using algorithms such as value iteration and policy gradient.\ This enables DRL to learn complex behavioral strategies without explicit supervision, which is especially suitable for tasks requiring long-term planning. However, traditional DRLs often face problems of low sample efficiency and unstable training when dealing with diverse and dynamically complex environments.\ 2024 MAMBA technology adopts a more efficient multi-objective optimization approach by improving the collaboration and gaming strategies among multiple intelligences, which not only enhances the coordination of the multi-intelligent systems, but also significantly improves the decision-making efficiency and stability.
	\par
    Compared with traditional DRL, Multi-agent Adversarial Reinforcement Learning (MAARL), as a new learning framework, is especially suitable for scenarios where multiple intelligences interact \cite{2017Mao}. In MAARL, each intelligent body not only optimizes its own behavioral strategies, but also needs to consider the behaviors and strategies of other intelligent bodies, thus forming a game-like learning process. Through adversarial learning, intelligences are able to evolve more robust decision-making strategies in their interactions with other intelligences. Although MAARL performs well in certain adversarial scenarios, it also faces the challenges of unstable training and poor convergence. In order to improve its stability, many improvements, such as Imitation Learning (ILL)-based approaches, have been proposed to further advance the application of this technique.
	\par
    Another technology, Large Scale Incremental Learning, breaks through the bottleneck of traditional learning methods when dealing with massive amounts of data. While traditional methods usually rely on batch learning, Large Scale Incremental Learning makes the learning process more flexible and real-time by gradually updating the model on the basis of streaming data. The method not only responds quickly to changes in new data during training, but also avoids overfitting through an effective storage mechanism, which improves the generalization ability in long-term learning tasks. The incremental learning-based framework can gradually adapt to new tasks in different environments, enabling the intelligences to maintain efficient learning and decision-making capabilities in dynamic and uncertain environments.
	\par
    Advances in self-supervised learning have also provided new ideas for the implementation of learning and evolutionary frameworks \cite{2020Jing}.\ The Forward-Forward algorithm, proposed in 2023, provides an innovative training method for self-supervised learning by predicting future states for model training. This approach avoids the reliance on large amounts of labeled data and drives the self-improvement of the learning process by generating future states. Compared to traditional supervised learning methods, self-supervised learning can utilize unlabeled data more efficiently and improve the generalization ability of the model. Its adaptive ability in dealing with unknown and complex environments is especially suitable for data-scarce scenarios in reinforcement learning and multi-intelligence systems.
	\par
    Together, these innovative approaches have contributed to the continuous improvement of learning and evolutionary frameworks, especially in applications in large-scale dynamic environments, where the learning ability of intelligences has been dramatically improved. From deep reinforcement learning to multi-intelligence games, incremental learning, and self-supervised learning, new algorithms continue to optimize the shortcomings of traditional frameworks, making intelligences more flexible and efficient when facing complex tasks.

\subsection{Action Modules}
The decision module performs task planning and sets up a specific sequence of actions to be passed to the action module, which executes the sequence sequentially or synchronously. The action module ensures that the actuators (e.g. robotic arms, motors, servos, etc.) can successfully complete the target task through precise motion control; at the same time, the feedback module optimizes the quality of the action through real-time acquisition of the execution feedback and closed-loop control to ensure the accuracy and efficiency of the task execution. Therefore, the main tasks of the action module can be categorized into motion control and feedback adjustment, and specific examples of innovation are shown in Figure~\ref{Fig6}.

\begin{figure}[htbp]
\centering
\includegraphics[width=0.9\linewidth]{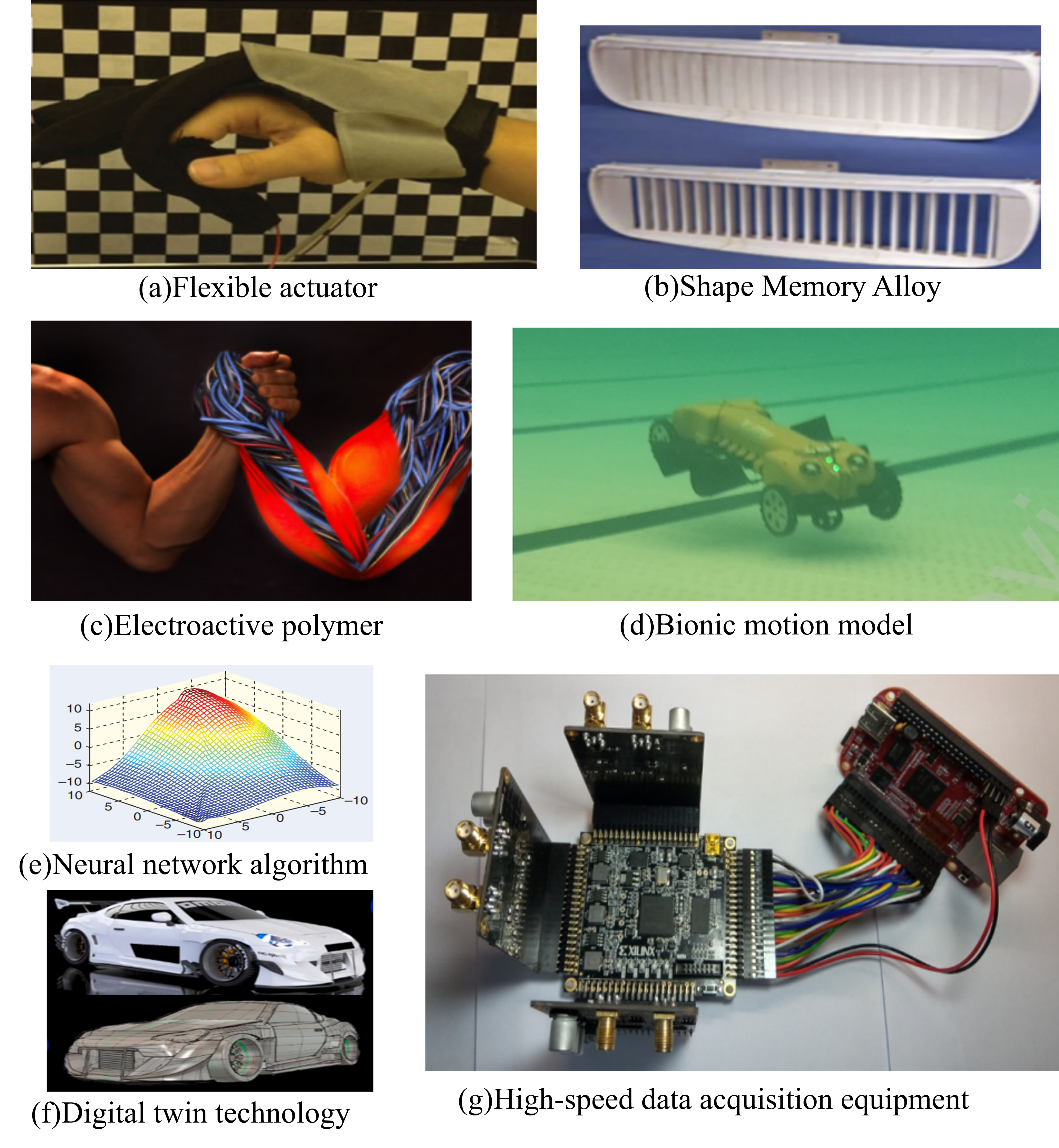}
\caption{The main tasks of the action module are divided into motion control and feedback adjustment. For motion control (a) Fig. \cite{2016Yeo} flexible sensors make the intelligent body's movements more natural and flexible with the help of materials such as (b) Fig. \cite{2014Jani} shape memory alloys and (c) Fig. \cite{SPIEBar-Cohen} electroactive polymers. Also (d) Fig. \cite{2024Chen} bionic motion modeling improves the ability to handle unstructured tasks. In terms of feedback adjustment, the combination of (e)Fig. \cite{2009Wilamowski} neural network algorithm and (g)Fig. \cite{2021Bhatti} \cite{lee2012renewable}high-speed data acquisition equipment improves the feedback response speed, and (f)Fig. \cite{2019Gu} digital twin technology is also invoked to improve the overall success rate of the task\cite{winkler2014security}.}\label{Fig6}
\end{figure}

	In motion control, the application of flexible actuators and smart materials has become a major highlight. With the help of materials such as shape memory alloys (SMA) and electroactive polymers (EAP), these technologies make the movements of intelligent bodies more natural and flexible, thus breaking through the limitations of traditional rigid design \cite{2018George}. In addition, high-precision servo control technology based on deep learning has been widely used. By adjusting the actuator parameters in real time, this technology can effectively enhance the adaptability of tasks in complex environments \cite{2017Castelli}. At the same time, the study of bionic motion modeling enables the intelligent body to mimic the biological multi-degree-of-freedom motion path, which improves the processing capability of unstructured tasks \cite{2005Dordevic}. The principle of autonomy is is demonstrated in this process, and the intelligent body is able to autonomously optimize its movements based on real-time feedback to improve execution efficiency.
	\par
   For the feedback adjustment task,the closed-loop control system realizes higher precision feedback adjustment \cite{2022Duan}. For example, in the grasping task, real-time sensing data was utilized to dynamically adjust the grasping force, reducing the task failure rate \cite{IEEEZhang}. Meanwhile, the combination of high-speed data acquisition equipment and neural network algorithms also improves the feedback response speed, enabling the intelligent body to dynamically optimize the motion path \cite{2023Righettini}. In addition, the introduction of digital twin technology allows the action module to predict possible execution problems through virtual models and optimize the solution in advance, thus improving the overall success rate of the task \cite{2022Botín-Sanabria}. The continuous innovation of these technologies also highlights the principle of efficiency optimization, which enables intelligent systems to not only perform tasks accurately, but also predict and adjust through virtual simulation to ensure optimal performance in changing environments.
\par
    The autonomous optimization capability and real-time feedback mechanism of the action module enable it to continuously learn and adapt to new tasks, which is highly compatible with the principle of “focus on potential, not deployment”. Intelligent bodies do not need to be pre-deployed in specific environments, but rather learn and adapt in real time to gradually improve their task execution capabilities. This capability not only enables intelligences to excel in complex and dynamic environments, but also lays a solid foundation for realizing AGI in a wider range of application scenarios.

\subsection{Feedback Modules}

The feedback information of the feedback module has been mentioned several times in the previous modules, and is an important part of the embodied intelligence system to achieve closed-loop control and self-optimization \cite{20Ren24}. It transforms data into feedback signals for adjusting the execution strategy of the action module by monitoring the interaction state of the intelligent body with the environment in real time, ensuring the accuracy and robustness of the task \cite{2021Yamakawa}. In this paper, the modules interacting with the feedback module are categorized from the feedback module, and the specific working principles of the feedback module and the perception module, the decision module, and the action module, respectively, will be explained next, as shown in Figure~\ref{Fig7}.

\begin{figure}[htbp]
\centering
\includegraphics[width=0.9\linewidth]{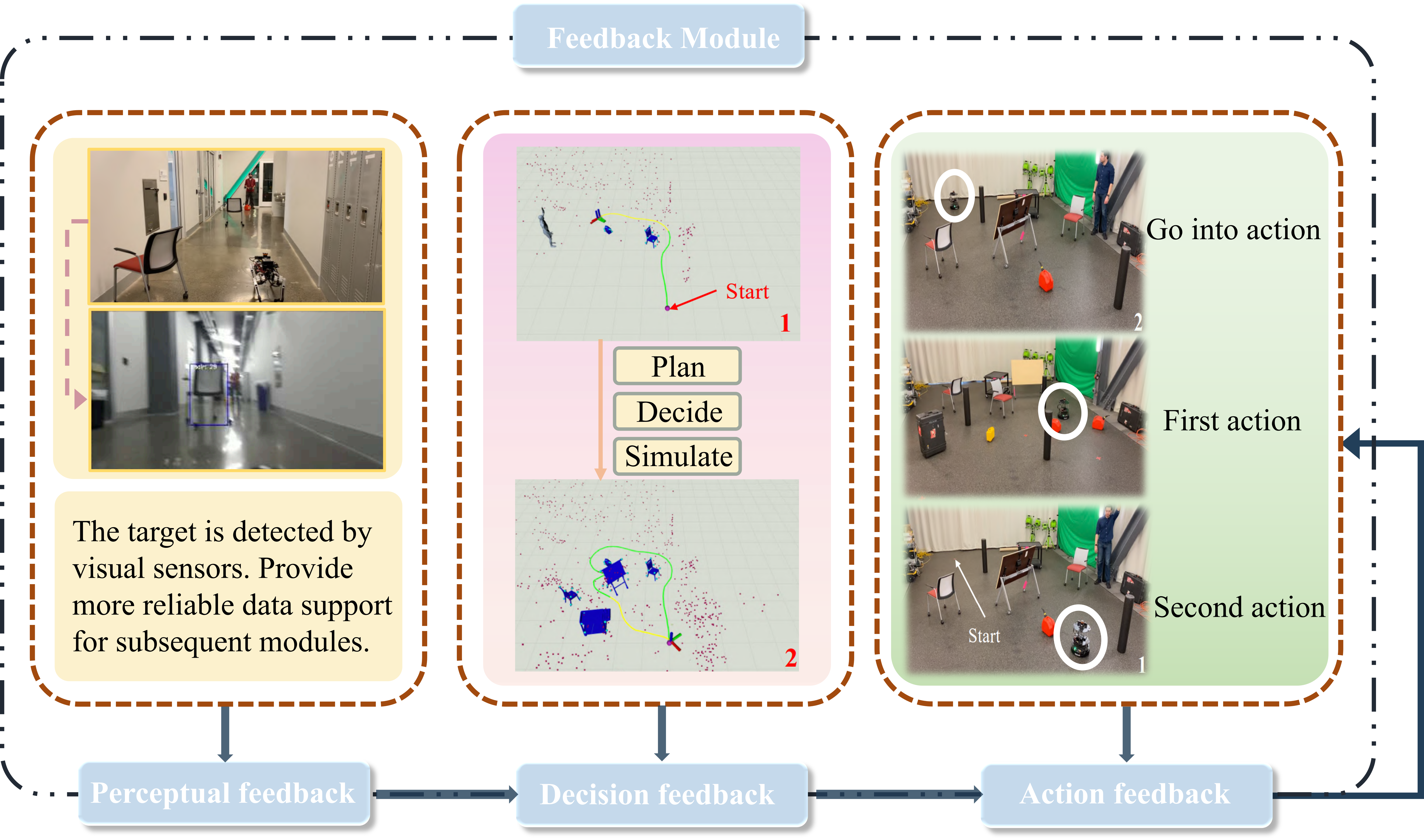}
\caption{The feedback module \cite{2020Vasilopoulos} is mainly divided into three main categories: perception feedback, decision-making feedback, and action feedback. Perception feedback dynamically adjusts the perception process by collecting external sensor data in real time. Decision feedback optimizes strategies and task settings by monitoring the actual effects of planning and decision-making. Action feedback is centered on the execution phase, and ensures the accuracy and efficiency of actions by monitoring and adjusting the actuator status in real time.}\label{Fig7}
\end{figure}

    Perceptual feedback is mainly responsible for the information interaction between the system and the environment, and dynamically adjusts the perception process by collecting external sensor data in real time \cite{1989Luo}. Specifically, static perceptual feedback focuses on capturing the characteristics of fixed environments or objects, such as the recognition of spatial layout, object shape, and material information, while dynamic perceptual feedback copes with rapidly changing scenarios, such as the localization and tracking of moving targets \cite{2006Schmidt}.\ Through these feedback mechanisms, the perception module is able to gradually improve the accuracy of environment understanding during task execution and provide more reliable data support for subsequent modules. For example, autonomous vehicles optimize speed adjustment and obstacle avoidance decisions through dynamic feedback when collecting road conditions in real time.
	\par
    Decision feedback optimizes strategy and task settings by monitoring the actual effects of planning and decision making \cite{2016Dennis}. Task-level feedback assesses the accomplishment of high-level task goals, e.g., determining whether a goal path reaches the desired endpoint, while strategy-level feedback focuses on the strengths and weaknesses of behavioral strategies, e.g., adapting a strategy to cope with unexpected environmental changes \cite{IEEECao}.\ Through this dual feedback mechanism, the decision module can correct planning errors and improve overall efficiency during task execution. For example, in logistics robots, task-level feedback is used to monitor the integrity of goods delivery, while policy-level feedback helps adjust the robot's navigation path in dynamic warehouse environments.
	\par
    Action feedback centers on the execution phase and ensures precise and efficient movements by monitoring and adjusting actuator states in real time \cite{2018Kress-Gazit}. Motion control feedback focuses on low-level actuator adjustment, e.g., closed-loop correction of angular deviation of the robotic arm \cite{2022Merckaert}; and execution effect feedback evaluates the final completion of the task action, e.g., determining whether the grasping task is successful or not. Action feedback is used throughout the task execution to ensure that the entire course of action remains consistent with expectations through real-time data closed-loop control.\ For example, industrial robots accurately complete assembly operations by monitoring the feedback of gripping position and force in real time, avoiding the accumulation of errors.
	\par
    The innovative technologies of the feedback module are mainly reflected in three aspects: real-time interaction, adaptive optimization and multimodal fusion feedback. Real-time feedback technology realizes rapid adjustment of execution actions through high-frequency sensors and closed-loop control algorithms, thus enhancing the immediacy and stability of task completion, which is especially suitable for dynamically changing high-complexity scenarios \cite{2014Ferrucci}. Adaptive optimization, on the other hand, makes use of machine learning methods such as reinforcement learning to enable the feedback mechanism to be dynamically adjusted according to the changes in the task and the environment, improving the execution efficiency and system robustness \cite{2021Gheibi}.In addition, multimodal feedback integrates multiple sensing information such as vision, haptics, and force sensing to provide richer environmental data for the perception module and optimize the task execution of the decision-making and action module through synergistic effects \cite{2020Huang}. The combination of these technologies makes the feedback module an important core of the embodied intelligent system, which enhances the system's adaptability to environmental changes and the precision of task execution. This characteristic of multimodal integration is highly compatible with the systemic principle of AGI, which is to achieve overall performance improvement by coordinating the synergistic work of different components within the system. At the same time, the adaptive optimization of the feedback module embodies the principle of continuous learning, which ensures that the system is able to improve and optimize its behavior over time based on experience, so that it is capable of handling more complex tasks and environmental changes.

\section{Future prospects and challenges}

\subsection{Future prospects}

The research and development of embodied intelligence is a key step toward AGI, and in the future, with the advancement of technology and the development of theory, embodied intelligence will face more opportunities and challenges in the process of transitioning to AGI. The following are some future perspectives of embodied intelligence in the transition to AGI.
	\par
    1. Improvement of adaptive and incremental learning: one important direction of embodied intelligence towards AGI is to improve the ability of adaptive and incremental learning. Currently, many embodied intelligence systems rely heavily on reinforcement learning to interact with the environment and obtain feedback. However, AGI systems need to be able to not only learn from every trial and error, but also to be able to migrate knowledge and skills across multiple different tasks. Through techniques such as meta-learning and lifelong learning, embodied intelligence is expected to become more flexible and efficient.\ AGI needs to be able to quickly adapt and demonstrate human-like migration capabilities when facing new environments or tasks. And embodied intelligence systems can gradually overcome the limitations of a single task through multi-task learning or multi-modal perception, and expand to more complex and broader intelligent tasks. This enhancement of capability will enable embodied intelligence to move towards AGI, so that it can not only learn new tasks, but also gradually accumulate experiences and migrate these experiences to different contexts.
	\par
    2. Fusion of cross-modal perception and cognition: when embodied intelligence moves towards AGI, the fusion of cross-modal perception and cognition will become the key to break through the bottleneck.\ AGI systems need to be able to process multiple sensory inputs, such as visual, tactile, auditory, etc., and to make effective reasoning, decision-making, and actions based on these perceptions. Currently, although embodied intelligence is able to learn and act through sensory information, the integration and processing of multimodal information remains a challenge.	Embodied intelligence may in the future integrate multiple sensory information such as vision, hearing and touch with the help of more advanced perceptual models and neural network architectures (e.g., graph neural networks, transformer models, etc.) to mimic the perceptual systems of human beings or other animals in order to gain a more comprehensive understanding of the environment. This cross-modal cognitive capability will provide strong support for the realization of AGI, enabling intelligences to cope with more complex tasks and environments.
	\par
    3. More efficient planning and decision-making capabilities: an important direction for embodied intelligence toward AGI is the development of capabilities that enable long-term planning and efficient decision-making. AGI requires intelligences to be able to make not only immediate decisions based on current perceptions, but also complex planning, reasoning, and multistep decision-making.\ Embodied intelligence systems must be able to plan themselves effectively in highly dynamic environments and adapt their decision-making strategies to external changes. Future embodied intelligence may achieve this goal through more advanced planning algorithms (e.g., reinforcement learning-based planning methods, recurrent neural network decision-making mechanisms, etc.). For example, AGI needs to have the ability to perform complex task decomposition, optimal resource allocation, and multi-objective planning, which will greatly enhance the autonomy and flexibility of embodied intelligence.
	\par
    4. Combination of Emotional and Social Intelligence: The combination of emotional and social intelligence will become another important direction in the development of embodied intelligence towards AGI, which not only needs to have problem-solving ability, but also needs to show a high degree of adaptability in human-computer interactions, teamwork and other social situations. Embodied intelligence has the potential to make intelligences behave more humanely in their interactions with humans or other intelligences in the future by learning social behaviors and emotional responses. For example, embodied intelligence can make appropriate responses and decisions by analyzing and understanding information such as human emotional expressions, language styles, and body language.\ Such capabilities will enable embodied intelligence to better integrate and efficiently cooperate with humans in human society, gradually moving closer to AGI goals. The future of embodied intelligence toward AGI is full of hope and challenges. With technological breakthroughs in cross-modal perception, multi-task learning, adaptive decision-making, emotional and social intelligence, embodied intelligence is expected to gradually break through the existing limitations and move towards AGI with high autonomy, adaptability and innovation.

\subsection{Challenges }

Embodied intelligence is the ability of a system to learn and adapt through interaction with and perception of the environment, and this intelligence not only relies on algorithms and data processing capabilities, but also includes the body's motor and perceptual mechanisms. Currently, embodied intelligence faces several challenges in its development towards AGI. These challenges are analyzed in detail in the following and discussed in relation to the requirements of AGI.
	\par
    1. The problem of coordination between perception and action: one of the core aspects of embodied intelligence is the ability to synergize perception and action. Machines need to explore and perceive their environment through body movements, and at the same time feedback these perceptions to adjust their movements. Achieving this is crucial for developing systems with AGI capabilities. However, the complexity of this process makes feedback mechanisms between perception and action a challenge.\ AGI requires systems to have the ability to collect information from multiple sensors (e.g., visual, tactile, auditory, etc.) and make decisions based on it.\ Whereas embodied intelligence requires continuous updating of perceptual information through physical actions, this interactive feedback mechanism requires the system to be able to perform efficient computation and decision making in real-time complex environments. Therefore, how to optimize the coordination between perception and action and reduce the feedback delay and error will directly affect the performance of the AGI system .
    \par
	2. Multi-task Learning and Adaptability Enhancement: AGI systems need to be able to flexibly switch between many different tasks and learn effectively in unseen tasks. This implies that the system must have strong multi-task learning capabilities and be able to migrate knowledge learned from one task to another. The multitasking capability of embodied intelligence relies heavily on its flexible perception and action mechanisms.\ The goal of AGI is to be able to handle complex tasks ranging from motor to cognitive in diverse and dynamically changing environments. Therefore, embodied intelligence has to be able to adapt and learn quickly with limited experience and environmental changes, avoiding overdependence on one type of task or specific environmental conditions .
	\par
    3. Self-understanding and planning capabilities in complex environments: AGIs need to not only understand the current environment, but also have the ability to plan and reason for themselves. This means that embodied intelligence must be able to make decisions in highly dynamic, complex environments and be able to reflect on and revise its decision-making process.\ Embodied intelligence usually acquires knowledge based on interaction with the environment, however, the variability and complexity of the environment places higher demands on AGI systems.\ The goal of AGI is to be able to make rational decisions in the face of uncertainty, which requires embodied intelligence to have effective self-understanding and the ability to plan for the long term, which is essential for long-term decision making by the machine.
	\par
    4. Integration of Learning and Reasoning Capabilities: One of the core features of AGI is a high degree of learning and reasoning capabilities, capable of solving problems through a combination of reasoning and learning. Embodied intelligence systems tend to learn through repeated trial and error and environmental interactions, but this does not mean that they are capable of reasoning, i.e., solving never-before-encountered tasks through existing knowledge.\ The main problem facing embodied intelligence is how to combine the results of perception with the ability to reason in the abstract to make effective decisions. agi requires systems that not only “remember” experiences, but also “understand” them to reason out new information. information. This abstract reasoning ability requires more complex algorithms and cognitive mechanisms, and how to integrate perceptual learning and reasoning systems is the key to the development of embodied intelligence towards AGI.
	\par
    5. Realization of long-term memory and transfer learning: AGI requires the system to be able to remember previous experiences for a long time and transfer learning based on these experiences. In the development of embodied intelligence, the ability of transfer learning and long-term memory is one of the important challenges, especially in the face of completely new or unseen environments, the ability of the system to quickly adapt and get help from previous experiences is the core capability of AGI. While current embodied intelligence systems often rely on short-term memory to accomplish tasks, AGI requires machines to be able to learn over the long term and migrate to apply these learnings in different contexts. Embodied intelligence must be able to accumulate experience and develop effective memory structures over long periods of interaction with complex environments, which is critical for AGI systems because it directly affects the machine's ability to adapt and innovate. 

\section{Conclusion}
Based on embodied intelligence systems, intelligences are able to perceive, understand, and interact with a wide range of objects in the virtual and physical worlds, demonstrating their key role in realizing general artificial intelligence (AGI). This paper provides a systematic overview of embodied robots, simulation platforms, and four core modules-perception module, intelligent decision module, action module, and feedback module, and discusses in depth how these modules work together to promote the development of AGI.  For example, the multimodal fusion of the perception module embodies the principle of 'focusing on versatility and performance', the adaptive learning mechanism of the decision module fits the principle of “focusing on potential rather than deployment', and precise control of the action module and the real-time optimization capability of the feedback module embody the core ideas of 'focusing on ecological validity' and 'focusing on capability rather than process'. This paper provides theoretical support and technical references for further exploration in this area in the future and points out the direction for the realization of AGI.



\bibliographystyle{unsrt}
\bibliography{ref}

\begin{thebibliography}{100}

\bibitem{dou2023towards}
Fei Dou, Jin Ye, Geng Yuan, Qin Lu, Wei Niu, Haijian Sun, Le~Guan, Guoyu Lu, Gengchen Mai, Ninghao Liu, et~al.
\newblock Towards artificial general intelligence (agi) in the internet of things (iot): Opportunities and challenges.
\newblock {\em arXiv preprint arXiv:2309.07438}, 2023.

\bibitem{2021Roy}
N.~Roy, I.~Posner, T.~Barfoot, P.~Beaudoin, Y.~Bengio, J.~Bohg, et~al.
\newblock From machine learning to robotics: Challenges and opportunities for embodied intelligence.
\newblock {\em arXiv preprint arXiv:2110.15245}, 2021.

\bibitem{grossberg2020path}
Stephen Grossberg.
\newblock A path toward explainable ai and autonomous adaptive intelligence: deep learning, adaptive resonance, and models of perception, emotion, and action.
\newblock {\em Frontiers in neurorobotics}, 14:36, 2020.

\bibitem{cianchetti2021embodied}
Matteo Cianchetti.
\newblock Embodied intelligence in soft robotics through hardware multifunctionality.
\newblock {\em Frontiers in Robotics and AI}, 8:724056, 2021.

\bibitem{summaira2021recent}
Jabeen Summaira, Xi~Li, Amin~Muhammad Shoib, Songyuan Li, and Jabbar Abdul.
\newblock Recent advances and trends in multimodal deep learning: A review.
\newblock {\em arXiv preprint arXiv:2105.11087}, 2021.

\bibitem{2024Zhou}
Y.~Zhou, L.~Huang, Q.~Bu, J.~Zeng, T.~Li, H.~Qiu, et~al.
\newblock Embodied understanding of driving scenarios.
\newblock {\em arXiv preprint arXiv:2403.04593}, 2024.

\bibitem{2024Liu}
Y.~Liu, W.~Chen, Y.~Bai, J.~Luo, X.~Song, K.~Jiang, et~al.
\newblock Aligning cyber space with physical world: A comprehensive survey on embodied ai.
\newblock {\em arXiv preprint arXiv:2407.06886}, 2024.

\bibitem{bariah2024ai}
Lina Bariah and M{\'e}rouane Debbah.
\newblock Ai embodiment through 6g: Shaping the future of agi.
\newblock {\em IEEE Wireless Communications}, 2024.

\bibitem{1950Turing}
A.~M. Turing.
\newblock Computing machinery and intelligence.
\newblock {\em Mind}, 59(236):433, 1950.

\bibitem{arslan2024artificial}
Suayb~S Arslan.
\newblock Artificial human intelligence: The role of humans in the development of next generation ai.
\newblock {\em arXiv preprint arXiv:2409.16001}, 2024.

\bibitem{floreano2008bio}
Dario Floreano and Claudio Mattiussi.
\newblock {\em Bio-inspired artificial intelligence: theories, methods, and technologies}.
\newblock MIT press, 2008.

\bibitem{2012Adams}
S.~Adams, I.~Arel, J.~Bach, R.~Coop, R.~Furlan, B.~Goertzel, et~al.
\newblock Mapping the landscape of human-level artificial general intelligence.
\newblock {\em AI Magazine}, 33(1):25--42, 2012.

\bibitem{2021Silver}
D.~Silver, S.~Singh, D.~Precup, and R.~S. Sutton.
\newblock Reward is enough.
\newblock {\em Artificial Intelligence}, 299:103535, 2021.

\bibitem{2021Garvey}
S.~C. Garvey.
\newblock The “general problem solver” does not exist: Mortimer taube and the art of ai criticism.
\newblock {\em IEEE Annals of the History of Computing}, 43(1):60--73, 2021.

\bibitem{1983Moto-Oka}
T.~Moto-Oka.
\newblock Overview to the fifth generation computer system project.
\newblock In {\em Proceedings of the 10th Annual International Symposium on Computer Architecture}, pages 417--422, 1983.

\bibitem{2002Roland}
A.~Roland and P.~Shiman.
\newblock {\em Strategic Computing: DARPA and the Quest for Machine Intelligence, 1983-1993}.
\newblock MIT Press, 2002.

\bibitem{2014Goertzel}
B.~Goertzel.
\newblock Artificial general intelligence: Concept, state of the art, and future prospects.
\newblock {\em Journal of Artificial General Intelligence}, 5(1):1, 2014.

\bibitem{2020Sejnowski}
T.~J. Sejnowski.
\newblock The unreasonable effectiveness of deep learning in artificial intelligence.
\newblock {\em Proceedings of the National Academy of Sciences}, 117(48):30033--30038, 2020.

\bibitem{2024Ge}
Y.~Ge, W.~Hua, K.~Mei, J.~Tan, S.~Xu, Z.~Li, and Y.~Zhang.
\newblock Openagi: When llm meets domain experts.
\newblock {\em Advances in Neural Information Processing Systems}, 36, 2024.

\bibitem{2004Pfeifer}
R.~Pfeifer and F.~Iida.
\newblock Embodied artificial intelligence: Trends and challenges.
\newblock {\em Lecture Notes in Computer Science}, pages 1--26, 2004.

\bibitem{1991Brooks}
R.~A. Brooks.
\newblock Intelligence without representation.
\newblock {\em Artificial Intelligence}, 47(1-3):139--159, 1991.

\bibitem{2001Pfeifer}
R.~Pfeifer and C.~Scheier.
\newblock {\em Understanding Intelligence}.
\newblock MIT Press, 2001.

\bibitem{2005Smith}
L.~B. Smith.
\newblock Cognition as a dynamic system: Principles from embodiment.
\newblock {\em Developmental Review}, 25(3-4):278--298, 2005.

\bibitem{black2012embodied}
John~B Black, Ayelet Segal, Jonathan Vitale, and Cameron~L Fadjo.
\newblock Embodied cognition and learning environment design.
\newblock In {\em Theoretical foundations of learning environments}, pages 198--223. Routledge, 2012.

\bibitem{golledge2018environmental}
Reginald~G Golledge, Gary~T Moore, Ronald Briggs, Martin~T Cadwallader, Ann~S Devlin, David~L George, Georgia Zannaras, Aleira Kreimer, Alfred~J Nigl, Harold~D Fishbein, et~al.
\newblock Environmental cognition.
\newblock In {\em Environmental design research}, pages 182--260. Routledge, 2018.

\bibitem{manakitsa2024review}
Nikoleta Manakitsa, George~S Maraslidis, Lazaros Moysis, and George~F Fragulis.
\newblock A review of machine learning and deep learning for object detection, semantic segmentation, and human action recognition in machine and robotic vision.
\newblock {\em Technologies}, 12(2):15, 2024.

\bibitem{li5128520sensor}
Tianrun Li, Zhimiao Yan, Yinghua Chen, and Ting Tan.
\newblock In-sensor multisensory integrative perception.
\newblock {\em Available at SSRN 5128520}.

\bibitem{manuelli2020keypoints}
Lucas Manuelli, Yunzhu Li, Pete Florence, and Russ Tedrake.
\newblock Keypoints into the future: Self-supervised correspondence in model-based reinforcement learning.
\newblock {\em arXiv preprint arXiv:2009.05085}, 2020.

\bibitem{2023Morris}
M.~R. Morris, J.~Sohl-Dickstein, N.~Fiedel, T.~Warkentin, A.~Dafoe, A.~Faust, et~al.
\newblock Levels of agi: Operationalizing progress on the path to agi.
\newblock {\em arXiv preprint arXiv:2311.02462}, 2023.

\bibitem{2021Hegde}
H.~M. Hegde.
\newblock {\em Autonomous Path Traversal and Object Avoidance in Cars-AirSim Simulation}.
\newblock PhD thesis, California State University, Northridge, 2021.

\bibitem{2014Nogueira}
L.~Nogueira.
\newblock Comparative analysis between gazebo and v-rep robotic simulators.
\newblock {\em Seminario Interno de Cognicao Artificial-SICA}, 2014(5):2, 2014.

\bibitem{2020Xia}
F.~Xia, W.~B. Shen, C.~Li, P.~Kasimbeg, M.~E. Tchapmi, A.~Toshev, et~al.
\newblock Interactive gibson benchmark (igibson 0.5): A benchmark for interactive navigation in cluttered environments.
\newblock {\em arXiv preprint arXiv:2005.04307}, 2020.

\bibitem{2021Coumans}
E.~Coumans and Y.~Bai.
\newblock Pybullet quickstart guide.
\newblock 2021.

\bibitem{2022Fernandez-Chaves}
D.~Fernandez-Chaves, J.~R. Ruiz-Sarmiento, A.~Jaenal, N.~Petkov, and J.~Gonzalez-Jimenez.
\newblock Robot@virtualhome, an ecosystem of virtual environments and tools for realistic indoor robotic simulation.
\newblock {\em Expert Systems with Applications}, 208:117970, 2022.

\bibitem{2007Ganapathy}
V.~Ganapathy and L.~W.~L. Dennis.
\newblock Design and implementation of a simulated robot using webots software.
\newblock In {\em Proceedings of the International Conference on Control, Instrumentation and Mechatronics Engineering}, pages 28--29, 2007.

\bibitem{coumans2021pybullet}
Erwin Coumans and Yunfei Bai.
\newblock Pybullet quickstart guide.
\newblock {\em ed: PyBullet Quickstart Guide. https://docs. google. com/document/u/1/d}, 2021.

\bibitem{2018Shigemi}
S.~Shigemi, A.~Goswami, and P.~Vadakkepat.
\newblock Asimo and humanoid robot research at honda.
\newblock {\em Humanoid Robotics: A Reference}, 55:90, 2018.

\bibitem{2020Liu}
H.~Liu, D.~Guo, F.~Sun, W.~Yang, S.~Furber, and T.~Sun.
\newblock Embodied tactile perception and learning.
\newblock {\em Brain Science Advances}, 6(2):132--158, 2020.

\bibitem{2024Qin}
Y.~Qin, E.~Zhou, Q.~Liu, Z.~Yin, L.~Sheng, R.~Zhang, et~al.
\newblock Mp5: A multi-modal open-ended embodied system in minecraft via active perception.
\newblock In {\em 2024 IEEE/CVF Conference on Computer Vision and Pattern Recognition (CVPR)}, pages 16307--16316, 2024.

\bibitem{2021Wang}
Y.~Wang.
\newblock Survey on deep multi-modal data analytics: Collaboration, rivalry, and fusion.
\newblock {\em ACM Transactions on Multimedia Computing, Communications, and Applications (TOMM)}, 17(1s):1--25, 2021.

\bibitem{2014Winkler}
T.~Winkler and B.~Rinner.
\newblock Security and privacy protection in visual sensor networks: A survey.
\newblock {\em ACM Computing Surveys (CSUR)}, 47(1):1--42, 2014.

\bibitem{2017Zou}
L.~Zou, C.~Ge, Z.~J. Wang, E.~Cretu, and X.~Li.
\newblock Novel tactile sensor technology and smart tactile sensing systems: A review.
\newblock {\em Sensors}, 17(11):2653, 2017.

\bibitem{2018Guo}
H.~Guo, X.~Pu, J.~Chen, Y.~Meng, M.~H. Yeh, G.~Liu, et~al.
\newblock A highly sensitive, self-powered triboelectric auditory sensor for social robotics and hearing aids.
\newblock {\em Science Robotics}, 3(20):eaat2516, 2018.

\bibitem{2019Shorten}
C.~Shorten and T.~M. Khoshgoftaar.
\newblock A survey on image data augmentation for deep learning.
\newblock {\em Journal of Big Data}, 6(1):1--48, 2019.

\bibitem{2002Elaksher}
A.~F. Elaksher and J.~S. Bethel.
\newblock Reconstructing 3d buildings from lidar data.
\newblock {\em International Archives of Photogrammetry Remote Sensing and Spatial Information Sciences}, 34(3/A):102--107, 2002.

\bibitem{2016García}
S.~García, S.~Ramírez-Gallego, J.~Luengo, J.~M. Benítez, and F.~Herrera.
\newblock Big data preprocessing: Methods and prospects.
\newblock {\em Big Data Analytics}, 1:1--22, 2016.

\bibitem{2017Wu}
J.~Wu.
\newblock Introduction to convolutional neural networks.
\newblock {\em National Key Lab for Novel Software Technology}, 2017.

\bibitem{2021Lindemann}
B.~Lindemann, T.~Müller, H.~Vietz, N.~Jazdi, and M.~Weyrich.
\newblock A survey on long short-term memory networks for time series prediction.
\newblock {\em Procedia CIRP}, 99:650--655, 2021.

\bibitem{2014Luo}
J.~Luo, W.~Wang, and H.~Qi.
\newblock Spatio-temporal feature extraction and representation for rgb-d human action recognition.
\newblock {\em Pattern Recognition Letters}, 50:139--148, 2014.

\bibitem{2022Huang}
K.~Huang, B.~Shi, X.~Li, X.~Li, S.~Huang, and Y.~Li.
\newblock Multi-modal sensor fusion for auto driving perception: A survey.
\newblock {\em arXiv preprint arXiv:2202.02703}, 2022.

\bibitem{2022Younis}
E.~M. Younis, S.~M. Zaki, E.~Kanjo, and E.~H. Houssein.
\newblock Evaluating ensemble learning methods for multi-modal emotion recognition using sensor data fusion.
\newblock {\em Sensors}, 22(15):5611, 2022.

\bibitem{2024Almujally}
N.~A. Almujally, A.~A. Rafique, N.~Al~Mudawi, A.~Alazeb, M.~Alonazi, A.~Algarni, et~al.
\newblock Multi-modal remote perception learning for object sensory data.
\newblock {\em Frontiers in Neurorobotics}, 18:1427786, 2024.

\bibitem{2015Simanek}
J.~Simanek, V.~Kubelka, and M.~Reinstein.
\newblock Improving multi-modal data fusion by anomaly detection.
\newblock {\em Autonomous Robots}, 39(2):139--154, 2015.

\bibitem{2020Bednarek}
M.~Bednarek, P.~Kicki, and K.~Walas.
\newblock On robustness of multi-modal fusion—robotics perspective.
\newblock {\em Electronics}, 9(7):1152, 2020.

\bibitem{2023Lin}
X.~Lin, S.~Chao, D.~Yan, L.~Guo, Y.~Liu, and L.~Li.
\newblock Multi-sensor data fusion method based on self-attention mechanism.
\newblock {\em Applied Sciences}, 13(21):11992, 2023.

\bibitem{2020Roheda}
S.~Roheda, H.~Krim, and B.~S. Riggan.
\newblock Robust multi-modal sensor fusion: An adversarial approach.
\newblock {\em IEEE Sensors Journal}, 21(2):1885--1896, 2020.

\bibitem{2018Nadon}
F.~Nadon, A.~J. Valencia, and P.~Payeur.
\newblock Multi-modal sensing and robotic manipulation of non-rigid objects: A survey.
\newblock {\em Robotics}, 7(4):74, 2018.

\bibitem{2009Noceti}
N.~Noceti, B.~Caputo, C.~Castellini, L.~Baldassarre, A.~Barla, L.~Rosasco, et~al.
\newblock Towards a theoretical framework for learning multi-modal patterns for embodied agents.
\newblock In {\em International Conference on Image Analysis and Processing}, pages 239--248, 2009.

\bibitem{2024Jia}
Z.~Jia, J.~Wang, and R.~Jin.
\newblock Grnet: A graph reasoning network for enhanced multi-modal learning in scene text recognition.
\newblock {\em The Computer Journal}, bxae085, 2024.

\bibitem{2022de}
J.~B. de~la Cita.
\newblock {\em Multimodal Perception for Autonomous Driving}.
\newblock PhD thesis, Universidad Carlos III de Madrid, 2022.

\bibitem{2003Fritsch}
J.~Fritsch, M.~Kleinehagenbrock, S.~Lang, T.~Plötz, G.~A. Fink, and G.~Sagerer.
\newblock Multi-modal anchoring for human–robot interaction.
\newblock {\em Robotics and Autonomous Systems}, 43(2-3):133--147, 2003.

\bibitem{2023Tang}
Q.~Tang, J.~Liang, and F.~Zhu.
\newblock A comparative review on multi-modal sensors fusion based on deep learning.
\newblock {\em Signal Processing}, page 109165, 2023.

\bibitem{2018Schwarting}
W.~Schwarting, J.~Alonso-Mora, and D.~Rus.
\newblock Planning and decision-making for autonomous vehicles.
\newblock {\em Annual Review of Control, Robotics, and Autonomous Systems}, 1(1):187--210, 2018.

\bibitem{2023Singh}
J.~Singh.
\newblock Advancements in ai-driven autonomous robotics: Leveraging deep learning for real-time decision making and object recognition.
\newblock {\em Journal of Artificial Intelligence Research and Applications}, 3(1):657--697, 2023.

\bibitem{2016Tsarouchi}
P.~Tsarouchi, S.~Makris, and G.~Chryssolouris.
\newblock Human–robot interaction review and challenges on task planning and programming.
\newblock {\em International Journal of Computer Integrated Manufacturing}, 29(8):916--931, 2016.

\bibitem{2000Dietterich}
T.~G. Dietterich.
\newblock Hierarchical reinforcement learning with the maxq value function decomposition.
\newblock {\em Journal of Artificial Intelligence Research}, 13:227--303, 2000.

\bibitem{2004Michel}
O.~Michel.
\newblock Cyberbotics ltd. webots™: Professional mobile robot simulation.
\newblock {\em International Journal of Advanced Robotic Systems}, 1(1):5, 2004.

\bibitem{2006Hohl}
L.~Hohl, R.~Tellez, O.~Michel, and A.~J. Ijspeert.
\newblock Aibo and webots: Simulation, wireless remote control and controller transfer.
\newblock {\em Robotics and Autonomous Systems}, 54(6):472--485, 2006.

\bibitem{2008Galindo}
C.~Galindo, J.~A. Fernández-Madrigal, J.~González, and A.~Saffiotti.
\newblock Robot task planning using semantic maps.
\newblock {\em Robotics and Autonomous Systems}, 56(11):955--966, 2008.

\bibitem{2006Alami}
R.~Alami, A.~Clodic, V.~Montreuil, E.~A. Sisbot, and R.~Chatila.
\newblock Toward human-aware robot task planning.
\newblock In {\em AAAI Spring Symposium: To Boldly Go Where No Human-Robot Team Has Gone Before}, pages 39--46, 2006.

\bibitem{2019Paxton}
C.~Paxton, Y.~Barnoy, K.~Katyal, R.~Arora, and G.~D. Hager.
\newblock Visual robot task planning.
\newblock In {\em 2019 International Conference on Robotics and Automation (ICRA)}, pages 8832--8838, 2019.

\bibitem{2021Gupta}
A.~Gupta, S.~Savarese, S.~Ganguli, and L.~Fei-Fei.
\newblock Embodied intelligence via learning and evolution.
\newblock {\em Nature Communications}, 12(1):5721, 2021.

\bibitem{2004Floreano}
D.~Floreano, F.~Mondada, A.~Perez-Uribe, and D.~Roggen.
\newblock Evolution of embodied intelligence.
\newblock In {\em Embodied Artificial Intelligence: International Seminar, Dagstuhl Castle, Germany, July 7-11, 2003. Revised Papers}, pages 293--311, 2004.

\bibitem{huang2021learning}
Junning Huang, Sirui Xie, Jiankai Sun, Qiurui Ma, Chunxiao Liu, Dahua Lin, and Bolei Zhou.
\newblock Learning a decision module by imitating driver’s control behaviors.
\newblock In {\em Conference on Robot Learning}, pages 1--10. PMLR, 2021.

\bibitem{2021Sadhu}
A.~Sadhu, T.~Gupta, M.~Yatskar, R.~Nevatia, and A.~Kembhavi.
\newblock Visual semantic role labeling for video understanding.
\newblock In {\em Proceedings of the IEEE/CVF Conference on Computer Vision and Pattern Recognition}, pages 5589--5600, 2021.

\bibitem{2023Guo}
H.~Guo, F.~Wu, Y.~Qin, R.~Li, K.~Li, and K.~Li.
\newblock Recent trends in task and motion planning for robotics: A survey.
\newblock {\em ACM Computing Surveys}, 55(13s):1--36, 2023.

\bibitem{1999Sutton}
R.~S. Sutton, D.~McAllester, S.~Singh, and Y.~Mansour.
\newblock Policy gradient methods for reinforcement learning with function approximation.
\newblock In {\em Advances in Neural Information Processing Systems}, volume~12, 1999.

\bibitem{2016Ho}
J.~Ho and S.~Ermon.
\newblock Generative adversarial imitation learning.
\newblock In {\em Advances in Neural Information Processing Systems}, volume~29, 2016.

\bibitem{2021Hospedales}
T.~Hospedales, A.~Antoniou, P.~Micaelli, and A.~Storkey.
\newblock Meta-learning in neural networks: A survey.
\newblock {\em IEEE Transactions on Pattern Analysis and Machine Intelligence}, 44(9):5149--5169, 2021.

\bibitem{2017Salimans}
T.~Salimans, J.~Ho, X.~Chen, S.~Sidor, and I.~Sutskever.
\newblock Evolution strategies as a scalable alternative to reinforcement learning.
\newblock {\em arXiv preprint arXiv:1703.03864}, 2017.

\bibitem{2023Hong}
X.~Hong, Y.~Lan, L.~Pang, J.~Guo, and X.~Cheng.
\newblock Visual reasoning: From state to transformation.
\newblock {\em IEEE Transactions on Pattern Analysis and Machine Intelligence}, 45(9):11352--11364, 2023.

\bibitem{2020Driess}
D.~Driess, J.~S. Ha, and M.~Toussaint.
\newblock Deep visual reasoning: Learning to predict action sequences for task and motion planning from an initial scene image.
\newblock {\em arXiv preprint arXiv:2006.05398}, 2020.

\bibitem{20Li24}
X.~Li, F.~Zhang, H.~Diao, Y.~Wang, X.~Wang, and L.~Y. Duan.
\newblock Densefusion-1m: Merging vision experts for comprehensive multimodal perception.
\newblock {\em arXiv preprint arXiv:2407.08303}, 2024.

\bibitem{2024Cai}
H.~Cai, Y.~Wang, L.~Liu, S.~Zhu, and M.~Chen.
\newblock Dagnn: Deep autoencoder-based graph neural network for local anomaly detection.
\newblock In {\em 2024 6th International Conference on Communications, Information System and Computer Engineering (CISCE)}, pages 988--994, 2024.

\bibitem{20Zhou24}
T.~Zhou, T.~Shi, H.~Gao, and W.~Rao.
\newblock Learning to optimize state estimation in multi-agent reinforcement learning-based collaborative detection.
\newblock {\em IEEE Transactions on Mobile Computing}, 2024.

\bibitem{2022Li}
L.~Li, T.~Zhou, W.~Wang, J.~Li, and Y.~Yang.
\newblock Deep hierarchical semantic segmentation.
\newblock In {\em Proceedings of the IEEE/CVF Conference on Computer Vision and Pattern Recognition}, pages 1246--1257, 2022.

\bibitem{2023Wang}
L.~Wang, X.~Wu, Y.~Zhang, X.~Zhang, L.~Xu, Z.~Wu, and A.~Fei.
\newblock Deepadain-net: Deep adaptive device-edge collaborative inference for augmented reality.
\newblock {\em IEEE Journal of Selected Topics in Signal Processing}, 2023.

\bibitem{2023Okubo}
T.~Okubo and M.~Takahashi.
\newblock Multi-agent action graph based task allocation and path planning considering changes in environment.
\newblock {\em IEEE Access}, 11:21160--21175, 2023.

\bibitem{2024Zhang}
J.~Zhang, J.~Ren, Y.~Cui, D.~Fu, and J.~Cong.
\newblock Multi-usv task planning method based on improved deep reinforcement learning.
\newblock {\em IEEE Internet of Things Journal}, 2024.

\bibitem{2024Liang}
Z.~Liang, Y.~Mu, H.~Ma, M.~Tomizuka, M.~Ding, and P.~Luo.
\newblock Skilldiffuser: Interpretable hierarchical planning via skill abstractions in diffusion-based task execution.
\newblock In {\em Proceedings of the IEEE/CVF Conference on Computer Vision and Pattern Recognition}, pages 16467--16476, 2024.

\bibitem{2024Klar}
M.~Klar, P.~Ruediger, M.~Schuermann, G.~T. Gören, M.~Glatt, B.~Ravani, and J.~C. Aurich.
\newblock Explainable generative design in manufacturing for reinforcement learning based factory layout planning.
\newblock {\em Journal of Manufacturing Systems}, 72:74--92, 2024.

\bibitem{2024Sun}
J.~Sun, Q.~Zhang, Y.~Duan, X.~Jiang, C.~Cheng, and R.~Xu.
\newblock Prompt, plan, perform: Llm-based humanoid control via quantized imitation learning.
\newblock In {\em 2024 IEEE International Conference on Robotics and Automation (ICRA)}, pages 16236--16242, 2024.

\bibitem{20Liang24}
Z.~Liang, R.~Yang, J.~Wang, L.~Liu, X.~Ma, and Z.~Zhu.
\newblock Dynamic constrained evolutionary optimization based on deep q-network.
\newblock {\em Expert Systems with Applications}, 249:123592, 2024.

\bibitem{20Liu24Liu}
Q.~Liu, Y.~Tian, T.~Zhou, K.~Lyu, R.~Xin, Y.~Shang, et~al.
\newblock A few-shot disease diagnosis decision making model based on meta-learning for general practice.
\newblock {\em Artificial Intelligence in Medicine}, 147:102718, 2024.

\bibitem{2024Ren}
J.~Ren, G.~Swamy, Z.~S. Wu, J.~A. Bagnell, and S.~Choudhury.
\newblock Hybrid inverse reinforcement learning.
\newblock {\em arXiv preprint arXiv:2402.08848}, 2024.

\bibitem{2024LiuJ}
J.~Liu, X.~Qi, P.~Hang, and J.~Sun.
\newblock Enhancing social decision-making of autonomous vehicles: A mixed-strategy game approach with interaction orientation identification.
\newblock {\em IEEE Transactions on Vehicular Technology}, 2024.

\bibitem{2024Sahoo}
G.~S. Sahoo, R.~H.~J. Rani, M.~K. Goyal, V.~A. Mohammed, D.~L.~F. Jana, and N.~N. Wasatkar.
\newblock Leveraging adaptive algorithms for real-time data analysis.
\newblock In {\em 2024 15th International Conference on Computing Communication and Networking Technologies (ICCCNT)}, pages 1--6, 2024.

\bibitem{2024Dong}
K.~Dong, Y.~Luo, Y.~Wang, Y.~Liu, C.~Qu, Q.~Zhang, et~al.
\newblock Dyna-style model-based reinforcement learning with model-free policy optimization.
\newblock {\em Knowledge-Based Systems}, 287:111428, 2024.

\bibitem{2024Shuford}
J.~Shuford.
\newblock Deep reinforcement learning unleashing the power of ai in decision-making.
\newblock {\em Journal of Artificial Intelligence General Science (JAIGS) ISSN: 3006-4023}, 1(1), 2024.

\bibitem{2024Rimon}
Z.~Rimon, T.~Jurgenson, O.~Krupnik, G.~Adler, and A.~Tamar.
\newblock Mamba: An effective world model approach for meta-reinforcement learning.
\newblock {\em arXiv preprint arXiv:2403.09859}, 2024.

\bibitem{2019Wachi}
A.~Wachi.
\newblock Failure-scenario maker for rule-based agent using multi-agent adversarial reinforcement learning and its application to autonomous driving.
\newblock {\em arXiv preprint arXiv:1903.10654}, 2019.

\bibitem{2019Wu}
Y.~Wu, Y.~Chen, L.~Wang, Y.~Ye, Z.~Liu, Y.~Guo, and Y.~Fu.
\newblock Large scale incremental learning.
\newblock In {\em Proceedings of the IEEE/CVF Conference on Computer Vision and Pattern Recognition}, pages 374--382, 2019.

\bibitem{2023Brenig}
J.~Brenig and R.~Timofte.
\newblock A study of forward-forward algorithm for self-supervised learning.
\newblock {\em arXiv preprint arXiv:2309.11955}, 2023.

\bibitem{2019Wang}
Z.~Wang, Y.~Wu, and Q.~Niu.
\newblock Multi-sensor fusion in automated driving: A survey.
\newblock {\em IEEE Access}, 8:2847--2868, 2019.

\bibitem{2016Cadena}
C.~Cadena, L.~Carlone, H.~Carrillo, Y.~Latif, D.~Scaramuzza, J.~Neira, et~al.
\newblock Past, present, and future of simultaneous localization and mapping: Toward the robust-perception age.
\newblock {\em IEEE Transactions on Robotics}, 32(6):1309--1332, 2016.

\bibitem{IEEEMulti-agent}
T.~Luo, B.~Subagdja, D.~Wang, and A.~H. Tan.
\newblock Multi-agent collaborative exploration through graph-based deep reinforcement learning.
\newblock In {\em 2019 IEEE International Conference on Agents (ICA)}, pages 2--7. IEEE, October 2019.

\bibitem{2023Multi-usv}
W.~Gan, X.~Qu, D.~Song, and P.~Yao.
\newblock Multi-usv cooperative chasing strategy based on obstacles assistance and deep reinforcement learning.
\newblock {\em IEEE Transactions on Automation Science and Engineering}, 2023.

\bibitem{2022Cooperative}
J.~Wang, Y.~Hong, J.~Wang, J.~Xu, Y.~Tang, Q.~L. Han, and J.~Kurths.
\newblock Cooperative and competitive multi-agent systems: From optimization to games.
\newblock {\em IEEE/CAA Journal of Automatica Sinica}, 9(5):763--783, 2022.

\bibitem{PMLRFan}
J.~Fan, Z.~Wang, Y.~Xie, and Z.~Yang.
\newblock A theoretical analysis of deep q-learning.
\newblock In {\em Learning for dynamics and control}, pages 486--489. PMLR, July 2020.

\bibitem{PMLRZintgraf}
L.~Zintgraf, K.~Shiarli, V.~Kurin, K.~Hofmann, and S.~Whiteson.
\newblock Fast context adaptation via meta-learning.
\newblock In {\em International Conference on Machine Learning}, pages 7693--7702. PMLR, May 2019.

\bibitem{2015Mnih}
V.~Mnih, K.~Kavukcuoglu, D.~Silver, A.~A. Rusu, J.~Veness, M.~G. Bellemare, et~al.
\newblock Human-level control through deep reinforcement learning.
\newblock {\em Nature}, 518(7540):529--533, 2015.

\bibitem{2017Mao}
H.~Mao, Z.~Gong, Y.~Ni, and Z.~Xiao.
\newblock Accnet: Actor-coordinator-critic net for "learning-to-communicate" with deep multi-agent reinforcement learning.
\newblock {\em arXiv preprint arXiv:1706.03235}, 2017.

\bibitem{2020Jing}
L.~Jing and Y.~Tian.
\newblock Self-supervised visual feature learning with deep neural networks: A survey.
\newblock {\em IEEE Transactions on Pattern Analysis and Machine Intelligence}, 43(11):4037--4058, 2020.

\bibitem{2016Yeo}
J.~C. Yeo, H.~K. Yap, W.~Xi, Z.~Wang, C.~H. Yeow, and C.~T. Lim.
\newblock Flexible and stretchable strain sensing actuator for wearable soft robotic applications.
\newblock {\em Advanced Materials Technologies}, 1(3):1600018, 2016.

\bibitem{2014Jani}
J.~M. Jani, M.~Leary, A.~Subic, and M.~A. Gibson.
\newblock A review of shape memory alloy research, applications and opportunities.
\newblock {\em Materials \& Design (1980-2015)}, 56:1078--1113, 2014.

\bibitem{SPIEBar-Cohen}
Y.~Bar-Cohen.
\newblock Electroactive polymers: current capabilities and challenges.
\newblock In {\em Smart Structures and Materials 2002: Electroactive Polymer Actuators and Devices (EAPAD)}, volume 4695, pages 1--7. SPIE, July 2002.

\bibitem{2024Chen}
L.~Chen, Q.~Hu, H.~Zhang, B.~Tong, X.~Shi, C.~Jiang, and L.~Sun.
\newblock Research on underwater motion modeling and closed-loop control of bionic undulating fin robot.
\newblock {\em Ocean Engineering}, 299:117400, 2024.

\bibitem{2009Wilamowski}
B.~M. Wilamowski.
\newblock Neural network architectures and learning algorithms.
\newblock {\em IEEE Industrial Electronics Magazine}, 3(4):56--63, 2009.

\bibitem{2021Bhatti}
G.~Bhatti, H.~Mohan, and R.~R. Singh.
\newblock Towards the future of smart electric vehicles: Digital twin technology.
\newblock {\em Renewable and Sustainable Energy Reviews}, 141:110801, 2021.

\bibitem{lee2012renewable}
Chang-Hun Lee.
\newblock Renewable and sustainable energy reviews.
\newblock {\em Coastal and Ocean}, 5(2):62--70, 2012.

\bibitem{2019Gu}
F.~C. Gu, H.~C. Chang, Y.~M. Hsueh, C.~C. Kuo, and B.~R. Chen.
\newblock Development of a high-speed data acquisition card for partial discharge measurement.
\newblock {\em IEEE Access}, 7:140312--140318, 2019.

\bibitem{winkler2014security}
Thomas Winkler and Bernhard Rinner.
\newblock Security and privacy protection in visual sensor networks: A survey.
\newblock {\em ACM Computing Surveys (CSUR)}, 47(1):1--42, 2014.

\bibitem{2018George}
T.~George~Thuruthel, Y.~Ansari, E.~Falotico, and C.~Laschi.
\newblock Control strategies for soft robotic manipulators: A survey.
\newblock {\em Soft Robotics}, 5(2):149--163, 2018.

\bibitem{2017Castelli}
F.~Castelli, S.~Michieletto, S.~Ghidoni, and E.~Pagello.
\newblock A machine learning-based visual servoing approach for fast robot control in industrial setting.
\newblock {\em International Journal of Advanced Robotic Systems}, 14(6):1729881417738884, 2017.

\bibitem{2005Dordevic}
G.~S. Dordevic, M.~Rasic, and R.~Shadmehr.
\newblock Parametric models for motion planning and control in biomimetic robotics.
\newblock {\em IEEE Transactions on Robotics}, 21(1):80--92, 2005.

\bibitem{2022Duan}
S.~Duan, Q.~Shi, and J.~Wu.
\newblock Multimodal sensors and ml-based data fusion for advanced robots.
\newblock {\em Advanced Intelligent Systems}, 4(12):2200213, 2022.

\bibitem{IEEEZhang}
J.~Zhang, C.~Song, Y.~Hu, and B.~Yu.
\newblock Improving robustness of robotic grasping by fusing multi-sensor.
\newblock In {\em 2012 IEEE International Conference on Multisensor Fusion and Integration for Intelligent Systems (MFI)}, pages 126--131. IEEE, September 2012.

\bibitem{2023Righettini}
P.~Righettini, R.~Strada, and F.~Cortinovis.
\newblock Neural network mapping of industrial robots’ task times for real-time process optimization.
\newblock {\em Robotics}, 12(5):143, 2023.

\bibitem{2022Botín-Sanabria}
D.~M. Botín-Sanabria, A.~S. Mihaita, R.~E. Peimbert-García, M.~A. Ramírez-Moreno, R.~A. Ramírez-Mendoza, and J.~D.~J. Lozoya-Santos.
\newblock Digital twin technology challenges and applications: A comprehensive review.
\newblock {\em Remote Sensing}, 14(6):1335, 2022.

\bibitem{20Ren24}
L.~Ren, J.~Dong, S.~Liu, L.~Zhang, and L.~Wang.
\newblock Embodied intelligence toward future smart manufacturing in the era of ai foundation model.
\newblock {\em IEEE/ASME Transactions on Mechatronics}, 2024.

\bibitem{2021Yamakawa}
Y.~Yamakawa, Y.~Matsui, and M.~Ishikawa.
\newblock Development of a real-time human-robot collaborative system based on 1 khz visual feedback control and its application to a peg-in-hole task.
\newblock {\em Sensors}, 21(2):663, 2021.

\bibitem{2020Vasilopoulos}
V.~Vasilopoulos, G.~Pavlakos, S.~L. Bowman, J.~D. Caporale, K.~Daniilidis, G.~J. Pappas, and D.~E. Koditschek.
\newblock Reactive semantic planning in unexplored semantic environments using deep perceptual feedback.
\newblock {\em IEEE Robotics and Automation Letters}, 5(3):4455--4462, 2020.

\bibitem{1989Luo}
R.~C. Luo and M.~G. Kay.
\newblock Multisensor integration and fusion in intelligent systems.
\newblock {\em IEEE Transactions on Systems, Man, and Cybernetics}, 19(5):901--931, 1989.

\bibitem{2006Schmidt}
P.~A. Schmidt, E.~Maël, and R.~P. Würtz.
\newblock A sensor for dynamic tactile information with applications in human–robot interaction and object exploration.
\newblock {\em Robotics and Autonomous Systems}, 54(12):1005--1014, 2006.

\bibitem{2016Dennis}
L.~A. Dennis, M.~Fisher, N.~K. Lincoln, A.~Lisitsa, and S.~M. Veres.
\newblock Practical verification of decision-making in agent-based autonomous systems.
\newblock {\em Automated Software Engineering}, 23:305--359, 2016.

\bibitem{IEEECao}
M.~Cao, A.~Stewart, and N.~E. Leonard.
\newblock Integrating human and robot decision-making dynamics with feedback: Models and convergence analysis.
\newblock In {\em 2008 47th IEEE Conference on Decision and Control}, pages 1127--1132. IEEE, December 2008.

\bibitem{2018Kress-Gazit}
H.~Kress-Gazit, M.~Lahijanian, and V.~Raman.
\newblock Synthesis for robots: Guarantees and feedback for robot behavior.
\newblock {\em Annual Review of Control, Robotics, and Autonomous Systems}, 1(1):211--236, 2018.

\bibitem{2022Merckaert}
K.~Merckaert, B.~Convens, C.~J. Wu, A.~Roncone, M.~M. Nicotra, and B.~Vanderborght.
\newblock Real-time motion control of robotic manipulators for safe human–robot coexistence.
\newblock {\em Robotics and Computer-Integrated Manufacturing}, 73:102223, 2022.

\bibitem{2014Ferrucci}
F.~Ferrucci and S.~Bock.
\newblock Real-time control of express pickup and delivery processes in a dynamic environment.
\newblock {\em Transportation Research Part B: Methodological}, 63:1--14, 2014.

\bibitem{2021Gheibi}
O.~Gheibi, D.~Weyns, and F.~Quin.
\newblock Applying machine learning in self-adaptive systems: A systematic literature review.
\newblock {\em ACM Transactions on Autonomous and Adaptive Systems (TAAS)}, 15(3):1--37, 2021.

\bibitem{2020Huang}
Z.~Huang, C.~Lv, Y.~Xing, and J.~Wu.
\newblock Multi-modal sensor fusion-based deep neural network for end-to-end autonomous driving with scene understanding.
\newblock {\em IEEE Sensors Journal}, 21(10):11781--11790, 2020.

\end{thebibliography}

\end{document}